%% file: main.tex
\documentclass{article}

\usepackage[main,final]{neurips_2026}

\usepackage[utf8]{inputenc}
\usepackage[T1]{fontenc}
\usepackage[hypertexnames=false]{hyperref}
\usepackage{url}
\usepackage{booktabs}
\usepackage{amsfonts}
\usepackage{amsmath}
\usepackage{amssymb}
\usepackage{nicefrac}
\usepackage{microtype}
\usepackage{xcolor}
\usepackage{graphicx}
\usepackage{colortbl}
\usepackage{multirow}
\usepackage{makecell}
\usepackage{tabularx}
\usepackage{enumitem}
\usepackage{algorithm}
\usepackage{algpseudocode}
\usepackage[most]{tcolorbox}
\usepackage{wrapfig}

\newcommand{\method}{MGMem}
\newcommand{\locomo}{LoCoMo}

\newtcolorbox{promptbox}[1]{
  enhanced, breakable,
  colback=white, colframe=black,
  colbacktitle=black, coltitle=white,
  title=#1, fonttitle=\bfseries,
  arc=4pt, boxrule=0.8pt,
  left=8pt, right=8pt, top=6pt, bottom=6pt,
  toptitle=4pt, bottomtitle=4pt
}

\title{MGMem: An Efficient, Deterministic, and Provenance-Preserving Framework for Long-Horizon Agent Memory}

\author{%
  Junhong Huang\textsuperscript{1} \quad Xin Tong\textsuperscript{1,2} \quad
  Jun Xia\textsuperscript{1,2}\thanks{Corresponding author.} \\
  \textsuperscript{1}The Hong Kong University of Science and Technology (Guangzhou) \\
  \textsuperscript{2}The Hong Kong University of Science and Technology \\
  \texttt{jhuang341@connect.hkust-gz.edu.cn} \quad
  \texttt{\{xint,junxia\}@hkust-gz.edu.cn}
}

\begin{document}
\maketitle

\begin{abstract}
Long-horizon Large Language Model (LLM) agents need persistent memory that surfaces the right past interaction at the right time across thousands of turns.
Many systems index by invoking generative LLMs to rewrite history into summaries, facts, or knowledge graphs---a step that is (a) costly (thousands of generative calls per build), (b) non-deterministic (undermining reproducibility and giving inconsistent answers across runs), and (c) provenance-erasing (the reader sees synthesized text, not the original turn).
We propose \method{}, a framework with no generative indexing calls that replaces this step with a deterministic discriminative-parser pipeline: recurring mentions across sessions are organized into an incidence graph, question mentions are resolved by a subject-conditioned three-stage resolver, and raw dialogue episodes are returned to the reader.
On English multi-session dialogue QA, \method{} outperforms matched-protocol baselines on LoCoMo without generative indexing calls; on the full LongMemEval-S, it remains competitive overall and leads or ties on single-session types among local retrieval comparisons. code: \url{https://github.com/junhong-h/MGMem}.
\end{abstract}

\section{Introduction}
\label{sec:intro}

\begin{wrapfigure}{r}{0.55\textwidth}
\vspace{-10pt}
\centering
\includegraphics[width=0.55\textwidth]{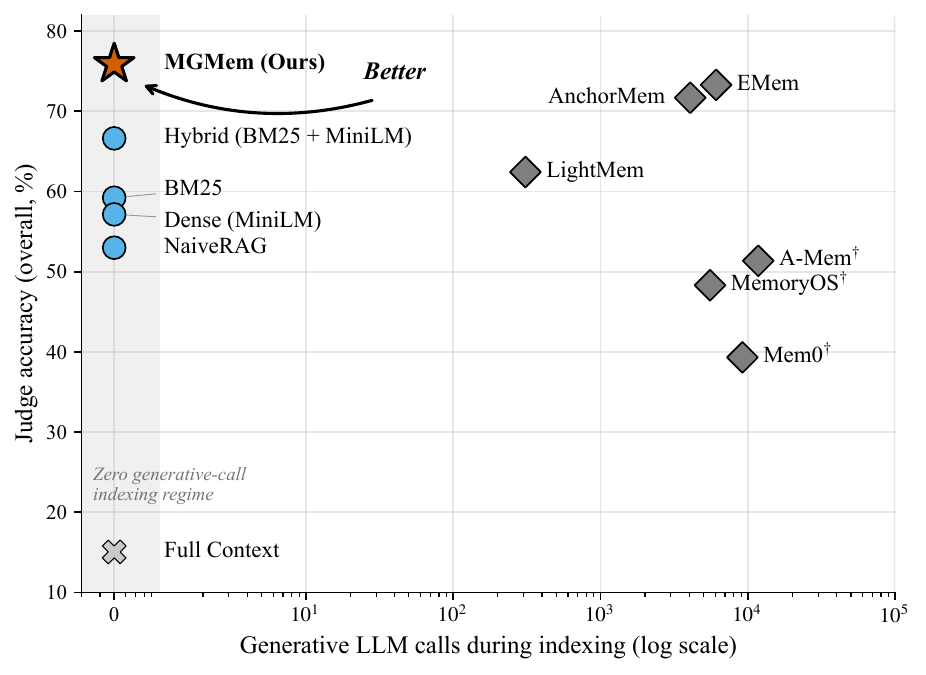}
\caption{Generative indexing calls vs.\ LoCoMo accuracy. Updated local results use three-run means; $^{\dagger}$ denotes quoted references (Section~\ref{sec:main-results}).}
\label{fig:pareto}
\vspace{-12pt}
\end{wrapfigure}

Long-horizon Large Language Model (LLM) agents---conversational companions, productivity assistants, multi-agent systems---are increasingly expected to remember interactions that span days, weeks, and months~\citep{packer2024memgpt,maharana2024locomo,wu2025longmemeval}.
Because language models are stateless beyond their context window, sustained interaction requires either ever-longer prompts---expensive, vulnerable to lost-in-the-middle effects that drop information from long contexts~\citep{liu2024lostinthemiddle,laban2025lostmultiturn}, and prone to LLM-side hallucination as context length stresses the model's capacity---or an external memory system that indexes past interactions and surfaces the right episode at query time.

\WFclear
The dominant design pattern for such indexed memory is \emph{generate-then-retrieve}: a generative LLM rewrites past interactions into summaries, facts, or knowledge graphs that the retriever then indexes~\citep{zhong2024memorybank,xu2025amem,kang2025memoryos,chhikara2025mem0,edge2024graphrag,gutierrez2025hipporag2}.
This places generative LLM inference on the indexing path: new interactions require extraction or rewriting calls, and the artifact that the reader ultimately consumes is no longer a turn from the original interaction.

This generative indexing step incurs three concrete drawbacks.
\textbf{(a) Cost.} Building memory for ten LoCoMo conversations requires $309$--$6{,}081$ generative indexing calls on average in our local runs of LightMem, AnchorMem, and EMem; complete build times range from $984$ to $2{,}816$ seconds (Table~\ref{tab:efficiency}).
\textbf{(b) Non-determinism.} Generative inference yields different summaries, facts, or graphs on repeated runs of the same input, so two indexing passes over the same dialogue may produce two different memories---and two different downstream answers, further undermining reproducibility, complicating debugging and ablation attribution, and giving end users an inconsistent agent across sessions.
\textbf{(c) Loss of provenance.} The reader consumes synthesized text rather than the original interaction; the trace from a returned answer back to the dialogue turn that supports it is broken or indirect.

We propose \method{}, a framework with no generative indexing calls for long-horizon agent memory that addresses all three drawbacks. \method{} replaces the generative indexing step with a deterministic discriminative-parser mention extractor, organizes recurring mentions across sessions into an incidence graph, and returns the original dialogue episodes to the reader so that the audit trail from answer to source is preserved by construction.
The framework is \emph{efficient} (zero generative LLM calls during indexing), \emph{deterministic} (the same input produces the same index), and \emph{provenance-preserving} (the reader sees raw dialogue episodes, never synthesized text).
Figure~\ref{fig:pareto} positions \method{} on the indexing-cost vs.\ end-to-end QA-accuracy plane, where it combines high accuracy with zero generative indexing calls.

\begin{figure*}[t]
\centering
\includegraphics[width=0.95\textwidth,height=0.3\textheight,keepaspectratio]{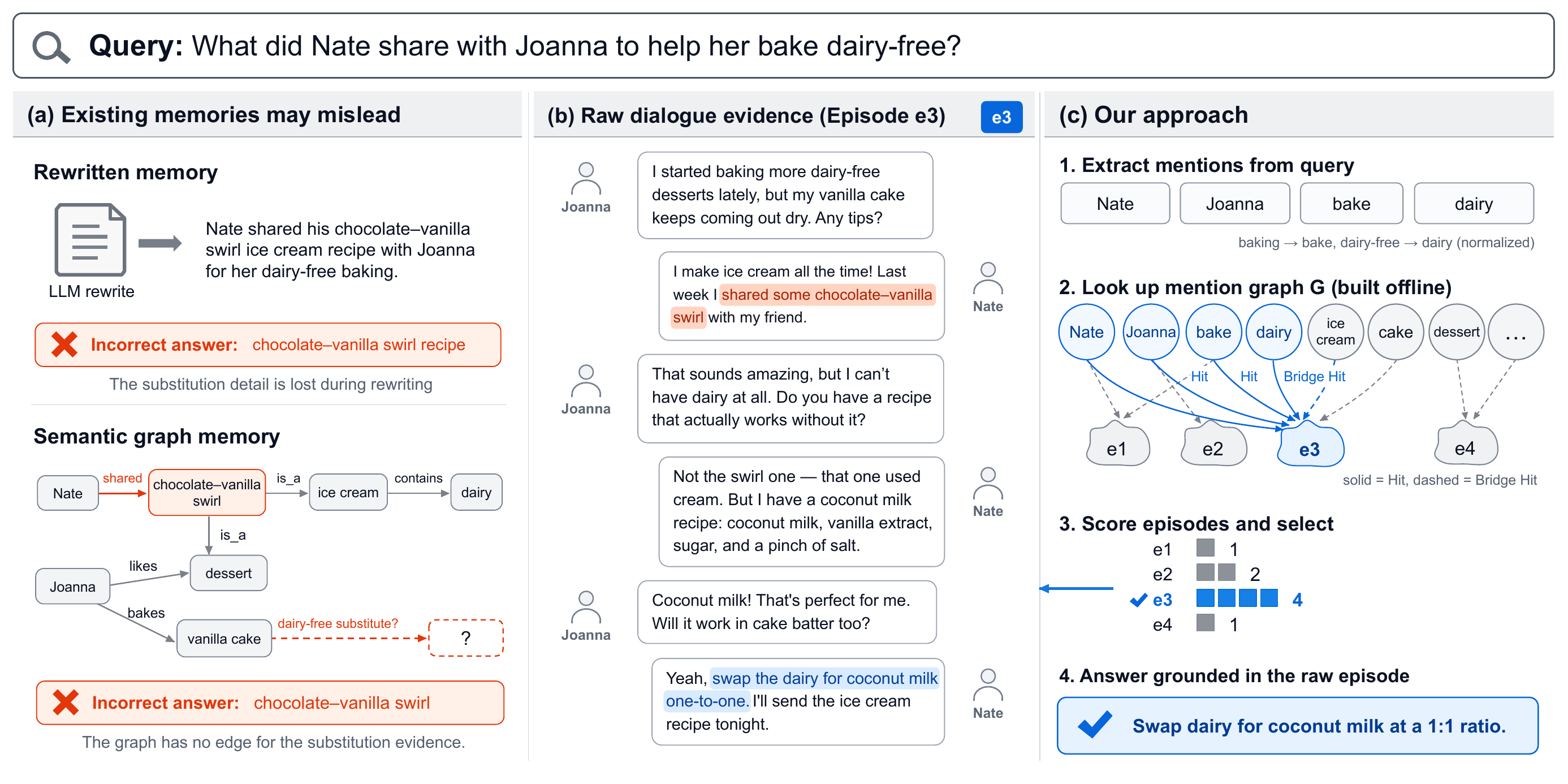}
\caption{An illustrative example of how three memory paradigms handle the same query.}
\label{fig:comparison}
\end{figure*}

\paragraph{Contributions.}
\begin{itemize}\setlength{\itemsep}{1pt}\setlength{\topsep}{2pt}
  \item[(1)] \method{} provides a memory-system architecture that is simultaneously efficient (zero generative indexing calls), deterministic, and provenance-preserving (raw dialogue episodes are returned to the reader).
  \item[(2)] We introduce a subject-conditioned three-stage resolver (exact $\to$ subject-local bridge $\to$ full-graph bridge) over an incidence graph of mention recurrence, with a stratified analysis pinpointing when the surface-recurrence prior is informative.
  \item[(3)] On English multi-session dialogue QA in LoCoMo, three locally matched runs yield $75.89\pm0.10\%$ accuracy ($+2.60$~pp over EMem; $+9.27$~pp over Hybrid) with zero generative indexing calls; on LongMemEval-S, \method{} remains competitive overall and leads or ties on single-session types among local retrieval comparisons (Appendix~\ref{appendix:generalization}).
\end{itemize}

\section{Related Work}

\paragraph{Long-horizon memory for LLM agents.}
Memory systems for multi-session dialogue differ in what they store and how aggressively they rewrite the interaction stream, with many adopting generative LLMs to produce the artifacts---notes, facts, summaries, hierarchical abstractions---that the reader consumes at query time.
A context-extension family pages or retrieves dialogue itself~\citep{packer2024memgpt,zhong2024memorybank}; a rewriting family compresses the stream into structured artifacts---A-Mem~\citep{xu2025amem} into evolving notes, Mem0~\citep{chhikara2025mem0} into compact facts and graph memories, MemoryOS~\citep{kang2025memoryos} and MemOS~\citep{li2025memos} into system-level memory tiers, LightMem~\citep{fang2025lightmem} into compressed summary streams; and a reference-indexing family explores finer units---AnchorMem~\citep{shen2026anchormem} atomic facts with preserved contexts, EMem~\citep{zhou2025emem} event-like EDUs, HyperMem~\citep{yue2026hypermem} hyperedges over topics, episodes, and facts.
For systems with LLM-generated indexing targets, cost, non-determinism, and indirect provenance enter during memory construction.

\paragraph{Graph- and IDF-based retrieval.}
A second line of work organizes retrieval through graph structure or surface-text statistics over a static external corpus: classical sparse retrieval grounded in Inverse Document Frequency (IDF) weighting~\citep{robertson2009bm25}, the Retrieval-Augmented Generation (RAG) family that pairs such retrievers with a downstream reader~\citep{gao2023ragsurvey}, and graph-based extensions such as GraphRAG~\citep{edge2024graphrag}, LightRAG~\citep{guo2025lightrag}, and HippoRAG~\citep{gutierrez2025hipporag2} that build entity or community graphs from documents and traverse them with personalized PageRank or graph summaries. These methods establish that surface and structural priors carry strong retrieval signal, but they operate over corpora that are external to the agent and stable across queries.

\paragraph{Positioning.}
\method{} draws on both lines: it adopts the IDF-style scoring and graph-traversal access patterns from the second, and targets the multi-session, user-specific interaction stream studied by the first.
The point of departure is the indexing target. Rather than organizing external documents by general semantic similarity or compressing dialogue through a generative LLM, \method{} indexes raw dialogue episodes by deterministic recurrence of user-specific mentions---a friend, a recurring activity, a project name---and returns the original episodes to the reader, so that the resulting memory is personalized by construction, reproducible, and directly traceable to the dialogue turns that support each answer.

\section{Method}
\label{sec:method}

\subsection{Problem Setup and Design Principles}

A long-horizon interaction stream consists of sessions $\mathcal{D} = \{S_1,\ldots,S_T\}$, partitioned into a set $\mathcal{E}$ of topical episodes (Section~\ref{sec:index}); we instantiate this on English multi-session dialogue. The formulation could in principle extend to assistant logs and other multi-session interaction records (untested; Section~\ref{sec:limitations}).
Given a query $q$, the memory system returns an ordered subset $\mathcal{C}_q \subseteq \mathcal{E}$ on which the reader conditions its answer.
Three design commitments instantiate the three advantages of \method{}.
\emph{Efficient}: the indexing pipeline must not call a generative LLM, with computation governed by parsing, graph construction, and retrieval (Appendix~\ref{appendix:graph-statistics}).
\emph{Deterministic}: every index component---nodes, edges, and the incidence relation---is a function of the surface text under fixed extractor and normalizer settings.
\emph{Provenance-preserving}: raw episodes are the source of truth; the memory locates them and never substitutes a generated artifact, so every retrieved episode has an audit trail back to specific dialogue turns; this does not guarantee faithful reader generation.
Concretely: nodes are surface handles, not LLM-generated propositions; edges encode observed episode-level co-occurrence, not entailment or causation.

\begin{figure*}[t]
\centering
\includegraphics[width=\linewidth]{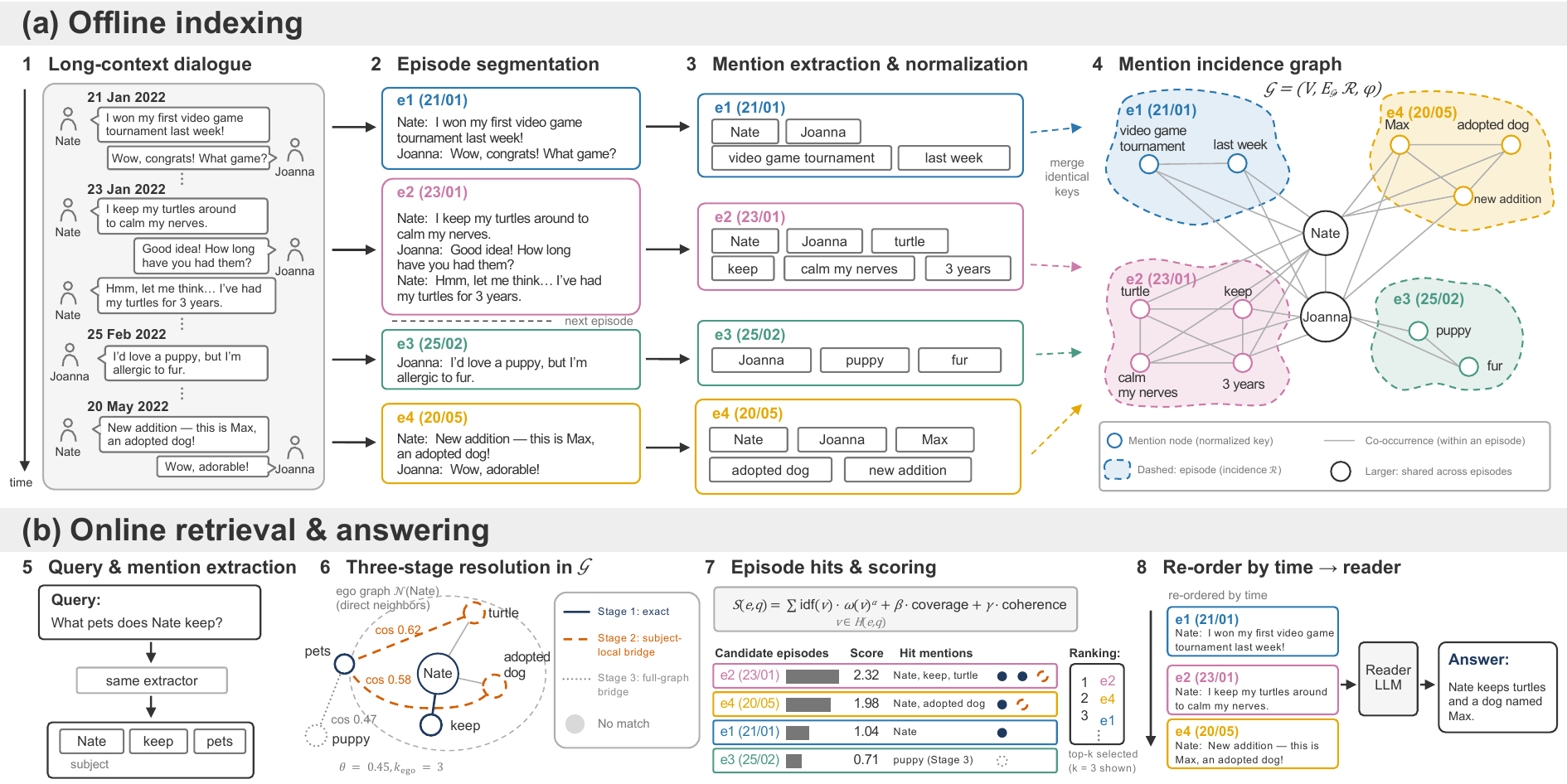}
\caption{Overview of the \method{} framework. Top: offline episodic indexing of the dialogue stream into the mention incidence graph $G$. Bottom: online retrieval resolves question mentions through the three-stage resolver, ranks candidate episodes, and returns raw dialogue to the reader.}
\label{fig:method-overview}
\end{figure*}

\subsection{Cross-Session Mention Incidence Graph}
\label{sec:graph}

\method{} represents long-term memory as an undirected weighted graph
\begin{equation}
G \;=\; (\mathcal{V},\, \mathcal{E}_G,\, \mathcal{R},\, \phi),
\end{equation}
where $\mathcal{V}$ is a set of normalized \emph{mention types}---people, places, objects, activities, dates, and other content-bearing expressions---each identified by a deterministic lexical key, so distinct surfaces such as \texttt{rainbow sidewalks} and \texttt{the rainbow sidewalk} collapse to a single node.
For each $e\in\mathcal{E}$, let $T_e\subseteq\mathcal{V}$ be its mention set.
Edges are weighted by episode-level co-occurrence,
\begin{equation}
\label{eq:cooccur}
w_{uv} \;=\; \bigl|\{e \in \mathcal{E} : u, v \in T_e\}\bigr|,
\end{equation}
and the incidence relation $\mathcal{R} = \{(v,e) : v \in T_e\}$ records every node's source episodes, giving $G$ its provenance: any node, edge, or downstream score can be traced back to specific dialogue turns.
Finally, $\phi(\cdot):\mathcal{V}\to\mathbb{R}^{d}$ is a dense embedding map over mention keys, used only for query-time bridging.
The graph is intentionally weaker than a knowledge graph: an edge records co-occurrence, not entailment or causation, and $\phi$ never rewrites dialogue content---this is what makes \method{} auditable.

\subsection{Deterministic Index Construction}
\label{sec:index}

\paragraph{Mention identification.}
Following the memory literature, ``LLM call'' denotes generative LLM inference; \method{} issues none during indexing.
We extract mentions with \texttt{spaCy}~\citep{honnibal2020spacy} using the transformer-based pipeline \texttt{en\_core\_web\_trf}, a discriminative parser that returns deterministic token-level annotations (POS, dependency, named entity).
A progressive cascade over the parse---named entities on a closed proper-noun whitelist, noun chunks, verb-headed subtrees, POS patterns, and content singletons---collects each turn's mentions in trust order; the union of a turn's mentions with the speaker identifier forms the episode mention set $T_e$. Cascade stages, the proper-noun whitelist, and a worked example are deferred to Appendix~\ref{appendix:reproducibility}.

\paragraph{Lexical normalization and merging.}
A normalizer $\nu(\cdot)$ maps each surface to a canonical key via speaker alias resolution if the surface matches a known speaker; otherwise by casefolding, stripping leading articles and punctuation, and per-token lemmatization. Two surfaces collapse iff $\nu(s_1)=\nu(s_2)$. The merge is deliberately lexical: \texttt{husband} and \texttt{spouse} stay apart at indexing time, since committing to a paraphrase before the query is known would no longer be reversible; semantic proximity is recovered at query time (Section~\ref{sec:resolve}).

\paragraph{Episode boundaries and incidence completion.}
Sessions are segmented at embedding change points (cohesion-drop peaks); a closed-world pass adds any literal-but-missed mention to $\mathcal{R}$ via word-boundary regex. Both passes are deterministic. Parameters and recomputation of $\mathrm{df}_v$, $\phi(v)$ are in Appendix~\ref{appendix:index-construction}.

\subsection{Subject-Conditioned Memory Resolution}
\label{sec:resolve}

The question $q$ is mapped to content mentions $M_q$ and subjects $A_q\subseteq M_q$ by running the \emph{same} extractor as indexing over the question text. A mention enters $A_q$ iff its dependency tag is \texttt{nsubj}/\texttt{nsubjpass} or it matches a known speaker alias. Source-aligned parsing is critical: with one extractor on both sides, modifier and verb tokens (e.g.\ \emph{specific themes}, \emph{explored}) become valid lookup keys rather than being silently dropped by a question-only LLM parser. Each $m\in M_q$ is then localized by a three-stage resolver, with at most $k{=}5$ accepted evidence nodes per mention (Stage~1 first; up to $k_{\mathrm{ego}}{=}3$ from Stage~2; remaining slots from Stage~3).

\textit{Stage~1 (exact).}
If $\nu(m) \in \mathcal{V}$, the corresponding node is taken with $\omega = 1.0$.

\textit{Stage~2 (subject-local bridge).}
Subjects are projected into the graph as $A_q^\star = \{\nu(a) : a \in A_q\} \cap \mathcal{V}$, defining the ego graph
\begin{equation}
\label{eq:ego}
N(A_q^\star) \;=\; \Bigl(\bigcup_{a \in A_q^\star} \{v \in \mathcal{V} : w_{av} > 0\}\Bigr) \setminus A_q^\star.
\end{equation}
For each $m$ unresolved by Stage~1, the top $k_{\mathrm{ego}}=3$ neighbors in $N(A_q^\star)$ above similarity floor $\theta = 0.45$ are accepted with $\omega = \cos(\phi(m), \phi(v))$.

\textit{Stage~3 (full-graph bridge).}
The remaining quota up to $k=5$ per mention is filled from $\mathcal{V}\setminus N(A_q^\star)$ above the same floor.

Stage~2 is the principal personalization mechanism (an abstract cue such as ``family'' is grounded in the queried subject's co-occurring mentions, not a global synonym pool); Stage~3's noisier provenance is propagated as lower confidence into the ranker.

\subsection{Episode Scoring and Selection}
\label{sec:rank}

Let $Z_q^{\mathcal{V}}$ be the set of evidence nodes from the resolver, with confidence $\omega(v)$ and a flag indicating an exact or bridged origin.
For each episode $e$, the hit set $H(e,q) = T_e \cap Z_q^{\mathcal{V}}$ collects matched nodes; episodes with $H = \varnothing$ are excluded.
The remaining episodes are ranked by an additive composition
\begin{equation}
\label{eq:score}
S(e,q) \;=\; S_{\mathrm{evi}}(e,q) + S_{\mathrm{cov}}(e,q) + S_{\mathrm{coh}}(e,q).
\end{equation}

The evidence term aggregates the IDF-weighted, confidence-modulated contribution of every hit:
\begin{equation}
\label{eq:s-evi}
S_{\mathrm{evi}}(e,q) \;=\; \sum_{v \in H(e,q)} \mathrm{idf}(v)\cdot \omega(v)^{\alpha},
\end{equation}
with $\omega(v) = 1$ for an exact (Stage~1) hit and $\omega(v) \in (0, 1)$ for a bridged hit (the $\cos$ similarity recorded by Stage~2 or 3).
The IDF is a square-root variant, $\mathrm{idf}(v) = \sqrt{\log(N/\mathrm{df}_v)}$ for $0 < \mathrm{df}_v < N$ (and zero for hub nodes with $\mathrm{df}_v = N$, such as the two speaker identifiers that appear in every episode), with $N = |\mathcal{E}|$. The $\sqrt{\cdot}$ compresses dynamic range so a single rare proper noun does not dominate scoring; the exponent $\alpha = 2$ amplifies the gap between an exact hit ($\omega=1$) and a low-confidence bridge ($\omega \approx 0.5$) by a factor of four, encoding the prior that one literal match outweighs several weak paraphrastic ones.

The coverage and coherence terms are
\begin{equation}
\label{eq:s-cov-coh}
\begin{aligned}
S_{\mathrm{cov}}(e,q)
&= \beta\,\frac{|H(e,q)|}{|Z_q^{\mathcal{V}}|}, \quad
S_{\mathrm{coh}}(e,q)
=
\begin{cases}
\gamma\,
\dfrac{\sum_{\{u,v\}\subseteq H(e,q)} w_{uv}}
{\binom{|H(e,q)|}{2} w_{\max}},
& |H(e,q)| \ge 2,\\
0, & |H(e,q)| < 2,
\end{cases}
\end{aligned}
\end{equation}
with production weights $\beta=0.3$, $\gamma=0.2$, and $w_{\max} = \max_{u\neq v} w_{uv}$. Both terms are bounded tie-shapers ($S_{\mathrm{cov}}\in[0,\beta]$, $S_{\mathrm{coh}}\in[0,\gamma]$) around the IDF-weighted evidence; the graph's deeper role is the incidence relation $\mathcal{R}$ plus subject-local ego expansion (Section~\ref{sec:resolve}), which together define how memory is accessed.

\paragraph{Selection, tie-breaking, and reader context.}
We rank the top-$10$ episodes for retrieval evaluation and pass the top-$5$ to the reader as raw dialogue (utterances with speaker labels, captions, and timestamps; no graph artifacts are surfaced). When two adjacent episodes' graph scores agree within numerical tolerance, a conservative dense tie-break consults whole-episode similarity to decide order; outside ties dense similarity is ignored. Tolerance value, dense similarity formulation, and the matched-protocol top-$k$ choice are documented in Appendix~\ref{appendix:reproducibility}.

\section{Experiments}
\label{sec:experiments}

\subsection{Settings}
\label{sec:settings}

\noindent\textbf{Datasets.}
We evaluate on \emph{two} long-term dialogue memory benchmarks.
The primary benchmark is \locomo{}~\citep{maharana2024locomo}, a multi-session conversational QA dataset with gold evidence annotations.
Following matched-protocol memory work~\citep{shen2026anchormem,fang2025lightmem}, we use the four-category subset (single-hop, multi-hop, temporal, open-domain), $1{,}540$ questions from ten conversations; the $446$ adversarial questions are evaluated separately (Appendix~\ref{appendix:additional}).
The second benchmark is LongMemEval-S~\citep{wu2025longmemeval}, an independently constructed long-horizon memory QA benchmark; we report results on the full $500$-question test set across six question types (single-session-user, single-session-assistant, single-session-preference, multi-session, temporal, knowledge-update) run end-to-end with the same reader and judge models as \locomo{}, with dataset-specific prompts and retrieval budgets detailed in Appendix~\ref{appendix:generalization} (Table~\ref{tab:longmemeval}).

\noindent\textbf{Baselines and baseline provenance.}
We compare against retrieval baselines (BM25~\citep{robertson2009bm25}, Dense, BM25+Dense Hybrid via reciprocal-rank fusion~\citep{cormack2009rrf}, NaiveRAG, HippoRAG~2~\citep{gutierrez2025hipporag2}) and generated-memory systems (A-Mem~\citep{xu2025amem}, Mem0~\citep{chhikara2025mem0}, LightMem~\citep{fang2025lightmem}, MemoryOS~\citep{kang2025memoryos}, AnchorMem~\citep{shen2026anchormem}, EMem~\citep{zhou2025emem}).
Dense and Hybrid use the \texttt{all-MiniLM-L6-v2} episode encoder~\citep{reimers2019sbert} consistent with prior \locomo{} memory comparisons.
\emph{MGMem, EMem, AnchorMem, LightMem, and Hybrid are locally rerun with seeds 42--44 under the same raw conversations, question set, reader, judge, QA prompt, and evaluation code.} Other generated-memory rows are retained as reference values from~\citet{shen2026anchormem,fang2025lightmem,xu2026structmem}; other local rows retain the original single-run results.
Primary comparisons use the locally matched reruns; literature values are contextual references rather than controlled local comparisons. We use the same judge prompt (Appendix~\ref{appendix:judge}).

\noindent\textbf{Implementation.}
The reader is GPT-4o-mini (\texttt{gpt-4o-mini-2024-07-18}) and the judge is \texttt{Qwen2.5-32B-Instruct}, both held fixed across methods; the separate graph-guidance ablation uses GPT-4o-mini as judge, while a Qwen2.5-32B-Instruct reader-substitution check is reported in Appendix~\ref{appendix:generalization} (Table~\ref{tab:qwen-main}). \method{} uses \texttt{spaCy} \texttt{en\_core\_web\_trf} as the source-aligned mention extractor on both dialogue and question sides, \texttt{all-MiniLM-L6-v2} for mention embeddings, and the production scoring configuration ($\sqrt{}$-IDF, $\alpha{=}2$, $\beta{=}0.3$, $\gamma{=}0.2$, $\theta{=}0.45$). Local reruns report mean $\pm$ sample standard deviation across seeds 42--44. The six retrieval settings were fixed on \texttt{conv-26}; its 152 answerable questions are also excluded in a separate 1,388-question analysis (Appendix~\ref{appendix:additional}); full model identifiers, hyperparameter neighborhoods, prompts, and trial pointers are in Appendices~\ref{appendix:reproducibility}, \ref{appendix:hyperparameters}, and \ref{appendix:judge}.

\noindent\textbf{Metrics.}
End-to-end QA accuracy is measured by an LLM-as-a-judge (primary) with token-level F1/BLEU-1 as secondary signals, on all $1{,}540$ four-category \locomo{} questions. Episode-level hit@$k$/MRR is reported on the $1{,}531$ questions with gold-episode annotations (the remaining $9$ unanchored queries are predominantly open-domain, with one each in multi-hop and temporal). Indexing cost includes complete wall time, generative calls/tokens, memory, and persisted index size.

\subsection{Main Results}
\label{sec:main-results}

Table~\ref{tab:main-qa} reports end-to-end QA accuracy, broken down by LoCoMo question category in addition to the overall column.
We organize the comparison into two parts: against generated-memory systems (which rewrite the dialogue into LLM-generated artifacts before indexing), and against retrieval baselines operating on raw dialogue units.

\paragraph{Versus generated-memory systems.}
\method{} reaches $\mathbf{75.89\pm0.10\%}$ overall judge accuracy, surpassing locally rerun EMem ($73.29\pm0.10\%$, $+2.60$~pp), AnchorMem ($71.69\pm0.36\%$, $+4.20$~pp), and LightMem ($62.42\pm0.10\%$, $+13.47$~pp). The paired MGMem--EMem conversation-bootstrap 95\% interval is $[+0.22,+5.15]$~pp; removing the development conversation preserves the ranking (Appendix~\ref{appendix:additional}).
The gap holds with zero generative indexing calls for \method{}; complete local indexing costs are reported in Section~\ref{sec:efficiency}.

\paragraph{Versus retrieval baselines.}
\method{} improves over the locally rerun Hybrid ($66.62\pm0.34\%$) by $9.27$~pp and has the largest mean accuracy in all four categories among the three-run comparisons. On open-domain, however, its mean accuracy of $49.65\%$ is below the $50.00\%$ reference results of Dense and HippoRAG~2. Original single-run retrieval rows remain reference results; Section~\ref{sec:stratification} analyzes the surface-recurrence prior separately.

\paragraph{Episode-level retrieval.}
\method{} reaches hit@1\,$=0.591$, hit@5\,$=0.837$, hit@10\,$=0.893$, MRR\,$=0.701$ (Table~\ref{tab:retrieval}), exceeding BM25, Dense, and Hybrid on every metric. Generated-memory systems retrieve synthesized artifacts whose units do not align with \locomo{} gold episodes, so episode-level metrics are restricted to raw-episode retrievers. The ``Full Context'' row in Table~\ref{tab:main-qa} feeds the entire dialogue with no retrieval; its $15.06\%$ accuracy reflects the lost-in-the-middle effect (Appendix~\ref{sec:reader-bottleneck}) and motivates indexed memory access.

\begin{table*}[t]
\centering
\footnotesize
\setlength{\tabcolsep}{4pt}
\renewcommand{\arraystretch}{1.15}
\begin{tabular}{@{}lccccc@{}}
\toprule
\textbf{Method} & \textbf{Single-Hop} & \textbf{Multi-Hop} & \textbf{Open-Domain} & \textbf{Temporal} & \textbf{Overall} \\
\midrule
\multicolumn{6}{l}{\textit{Retrieval baselines}} \\
\quad BM25                           & 71.11 & 34.75 & 38.54 & 55.76 & 59.22 \\
\quad Dense$^{\dagger}$              & 64.21 & 48.23 & 50.00 & 48.91 & 57.14 \\
\quad Hybrid$^{\ddagger}$ & $78.83$ & $46.45$ & $47.22$ & $58.15$ & $66.62$ \\
\quad NaiveRAG                       & 61.24 & 41.49 & 45.83 & 43.61 & 52.99 \\
\quad HippoRAG~2                     & 66.11 & 55.67 & 50.00 & 59.50 & 61.82 \\
\quad Full Context                   & 14.15 & 18.79 & 27.08 & 10.59 & 15.06 \\
\midrule
\multicolumn{6}{l}{\textit{Generated-memory systems}} \\
\quad A-Mem                          & 57.40 & 42.91 & 23.96 & 51.09 & 51.35 \\
\quad Mem0                           & 45.54 & 30.85 & 28.12 & 33.96 & 39.31 \\
\quad LightMem & $67.30$ & $51.30$ & $45.83$ & $64.38$ & $62.42$ \\
\quad MemoryOS                       & 55.65 & 56.74 & 42.71 & 23.36 & 48.31 \\
\quad AnchorMem & $81.21$ & $56.62$ & $48.26$ & $66.98$ & $71.69$ \\
\quad EMem & $85.77$ & $54.61$ & $46.18$ & $65.11$ & $73.29$ \\
\midrule
Ours & $\mathbf{87.79}$ & $\mathbf{57.80}$ & $\mathbf{49.65}$ & $\mathbf{68.43}$ & $\mathbf{75.89}$ \\
\bottomrule
\end{tabular}
\caption{LoCoMo QA accuracy (\%); reader: GPT-4o-mini, judge: Qwen2.5-32B-Instruct. MGMem, Hybrid, EMem, AnchorMem, and LightMem report means over three locally matched runs (seeds 42--44); bold marks the largest mean among these methods. Other retrieval rows retain original single-run values; A-Mem, Mem0, MemoryOS, and HippoRAG~2 are quoted references. $^{\dagger}$ MiniLM encoder; $^{\ddagger}$ BM25+Dense RRF ($k=60$).}
\label{tab:main-qa}
\end{table*}

\begin{table}[t]
\centering
\centering
\footnotesize
\setlength{\tabcolsep}{2pt}
\renewcommand{\arraystretch}{1.15}
\begin{tabular}{@{}lcccc@{}}
\toprule
\textbf{Method} & hit@1 & hit@5 & hit@10 & MRR \\
\midrule
BM25                          & 0.182 & 0.611 & 0.746 & 0.369 \\
Dense$^{\dagger}$             & 0.325 & 0.590 & 0.704 & 0.451 \\
Hybrid$^{\ddagger}$           & \underline{0.443} & \underline{0.706} & \underline{0.805} & \underline{0.564} \\
NaiveRAG                      & 0.249 & 0.550 & 0.678 & 0.392 \\
\midrule
\rowcolor{black!4}
\textbf{Ours}                 & \textbf{0.591} & \textbf{0.837} & \textbf{0.893} & \textbf{0.701} \\
\bottomrule
\end{tabular}
\caption{Episode-level retrieval quality; hit@$k$ and MRR as fractions in $[0,1]$. Best in \textbf{bold}, second-best \underline{underlined}. $^{\dagger}$\,\texttt{all-MiniLM-L6-v2} encoder; $^{\ddagger}$\,BM25 fused with Dense$^{\dagger}$ via reciprocal-rank fusion}
\label{tab:retrieval}

\end{table}

\begin{table}[t]
\centering\small
\setlength{\tabcolsep}{5pt}
\begin{tabular}{@{}lrrr@{}}
\toprule
Method & Complete indexing (s) & Gen. calls & Gen. tokens (M) \\
\midrule
\method{} & $199.02\pm43.37$ & 0 & 0 \\
AnchorMem & $984.26\pm41.76$ & $4{,}058.7\pm21.1$ & $3.989\pm0.029$ \\
EMem & $2{,}816.44\pm273.39$ & $6{,}081.3\pm5.5$ & $3.792\pm0.004$ \\
LightMem & $2{,}552.47\pm100.63$ & $309.0\pm4.4$ & $0.860\pm0.003$ \\
BM25 & $1.67\pm0.08$ & 0 & 0 \\
Dense & $13.62\pm0.17$ & 0 & 0 \\
Hybrid & $13.53\pm0.76$ & 0 & 0 \\
NaiveRAG & $12.98\pm0.39$ & 0 & 0 \\
\bottomrule
\end{tabular}
\caption{Complete indexing cost over all ten LoCoMo conversations, mean $\pm$ sample standard deviation across three fresh-index local runs. Wall time includes initialization, segmentation, parsing, embedding, graph construction, and serialization as applicable. Software installation and model downloads are excluded. Generative calls/tokens are separate from total computation.}
\label{tab:efficiency}
\end{table}

\subsection{Efficiency and Provenance}
\label{sec:efficiency}

Table~\ref{tab:efficiency} reports complete indexing costs over three fresh-index runs on the same Apple M5 Pro, from raw input to persisted retrieval-ready indices; installation and model downloads are excluded.
\method{} takes $199.02\pm43.37$~s, versus $984.26\pm41.76$~s for AnchorMem, $2{,}816.44\pm273.39$~s for EMem, and $2{,}552.47\pm100.63$~s for LightMem, with zero generative indexing calls/tokens. BM25, Dense, and Hybrid build faster than \method{}; zero generative calls do not imply zero computation.
The contrast is structural: every generated-memory system pays for an extraction or summarization step whose output becomes the storage layer the reader consumes; \method{} replaces that step with a deterministic discriminative parser plus a co-occurrence graph, removing the generative inference path during indexing entirely.

\noindent\textbf{Provenance.}
Because the reader sees raw dialogue episodes, retrieved evidence is traceable to specific turns through $\mathcal{R}$; generated answers can still be unsupported. A stratified 200-answer human audit finds $94.00\%$ evidence support and $94.50\%$ human--judge agreement (Appendix~\ref{appendix:additional}).

\noindent\textbf{Online cost.}
In the three-run profiling experiment, retrieval and reader wall totals over LoCoMo are $159.36\pm45.10$~s and $207.28\pm4.34$~s, respectively; resource details are in Appendix~\ref{appendix:reproducibility}.
The reader sees at most five raw episodes. This bounds the returned item count, while token count depends on segmentation and retrieval still depends on graph size.

\noindent\textbf{LongMemEval transfer.}
On the full LongMemEval-S benchmark ($500$ questions, six types), \method{} reaches $72.60\%$ overall, versus BM25 at $74.60\%$ and Hybrid at $73.80\%$,and leads or ties on single-session types among local retrieval comparisons (Table~\ref{tab:longmemeval}; quoted generated-memory rows are not strictly matched-protocol).

Additional ablations on the IDF variant and dense-fusion strategy, as well as sensitivity sweeps over the resolver hyperparameters $\theta$, $k$, and the scoring weights $\alpha,\beta,\gamma$, are reported in Appendices~\ref{appendix:additional-ablations}--\ref{appendix:hyperparameters}.

\section{Analysis}
\label{sec:analysis}

\subsection{When is the surface-recurrence prior informative?}
\label{sec:stratification}

To explain the category-dependent pattern in Section~\ref{sec:main-results}, we stratify the $1{,}540$ \locomo{} queries by (i) whether at least one non-subject content mention exact-matches a graph node, and (ii) the maximum IDF tertile of the query's content mentions; subjects and speakers are excluded so the stratification reflects content-reference recurrence.
This is the original single-run stratification, separate from the three-run main table. Figure~\ref{fig:strat-ablation}\,(a) reports the judge accuracy of \method{} and the Hybrid retriever in each bucket.
The gap is largest in the \emph{No exact match} bucket ($+13.10$~pp, $n{=}84$): when no query mention exact-matches a graph node, Stage 2/3 bridges recover evidence. This tests English query-side lexical mismatch, rather than a dedicated paraphrase benchmark.
Among exact-match buckets, the \emph{Medium}-IDF tertile leads ($+9.85$~pp, $n{=}485$) over the \emph{High}-IDF ($+7.96$~pp, $n{=}540$) and \emph{Low}-IDF ($+7.89$~pp, $n{=}431$) tertiles: High-IDF queries typically resolve on a single rare proper-noun match where Stage~1 alone suffices, while Medium-IDF queries combine multiple moderately rare mentions, which is where subject-local bridging contributes most.
Within LoCoMo's answerable categories, mention indexing provides gains across the analyzed surface-match and IDF buckets: the largest gain occurs when surface match is absent ($+13.1$~pp on \emph{No exact match}, where dense retrieval would normally be the fallback), and gains remain $+7.9$ to $+9.9$~pp across IDF tertiles when surface match is present, supporting its use as an always-on substrate in this LoCoMo setting.

\subsection{What carries the personalization gain?}
\label{sec:component-ablation}

\begin{figure}[t]
\centering
\begin{minipage}[t]{0.49\linewidth}
  \centering
  \includegraphics[width=\linewidth]{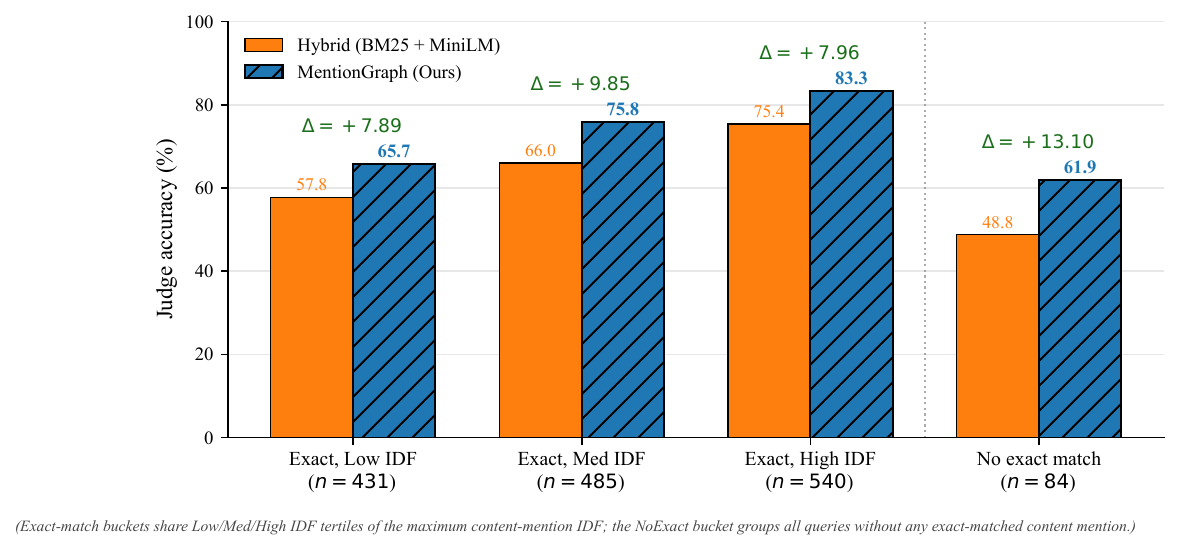}\\
  {\small (a) Stratification by exact-match $\times$ IDF tertile.}
\end{minipage}\hfill
\begin{minipage}[t]{0.49\linewidth}
  \centering
  \includegraphics[width=\linewidth]{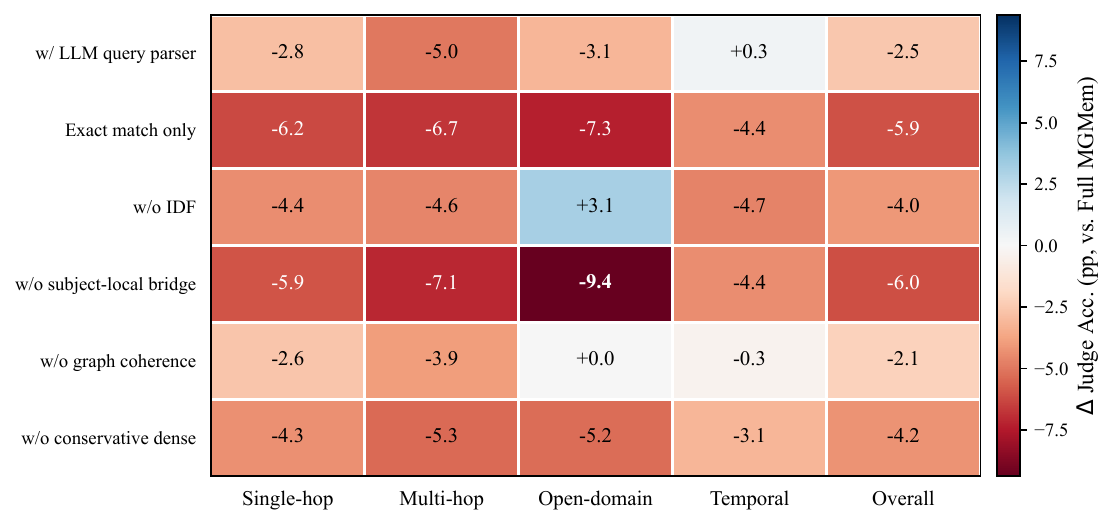}\\
  {\small (b) Component ablation by LoCoMo category ($\Delta$\,pp vs.\ Full).}
\end{minipage}
\caption{(a) Stratification of \locomo{} queries by exact-match $\times$ IDF tertile. Each bucket pairs Hybrid (orange) with \method{} (blue, hatched); $\Delta$ above the bar pair is the judge-accuracy gap in pp. The first three buckets cover queries with at least one exact-matched content mention split by Low / Med / High IDF tertile; the rightmost \emph{No exact} bucket groups queries with no exact-matched content mention. (b) Component ablation by \locomo{} category: each cell shows $\Delta$ judge accuracy in pp relative to Full \method{} (red\,$=$\,drop, blue\,$=$\,gain).}
\label{fig:strat-ablation}
\end{figure}

The following original single-run component analysis remains separate from the main reruns. Figure~\ref{fig:strat-ablation}\,(b) removes one mechanism at a time and breaks the judge-accuracy drop down by category; full retrieval metrics are in Appendix~\ref{appendix:additional-ablations}, Table~\ref{tab:ablation}.
\textbf{Subject-conditioned bridging is the largest lever.} Removing the subject-local bridge stage costs $-6.1$~pp ($-5.7$~pp hit@1, the largest single drop); removing all bridging (\emph{exact match only}) costs a near-identical $-6.0$~pp, while disabling only Stage~3 (full-graph bridge) costs $-0.26$~pp hit@1. Subject-local bridging carries the largest gain; Stage~3 provides residual recall when subject anchors are unavailable (Appendix~\ref{appendix:additional-ablations}).
\textbf{Source-aligned parsing and conservative dense are next.} Replacing the \texttt{spaCy} question parser with a separate LLM parser drops accuracy by $-2.6$~pp; conservative dense tie-breaking ($-4.3$~pp) and IDF weighting ($-4.1$~pp) contribute secondary lifts.
\textbf{Per-category.} Subject-local bridge removal has the deepest single-category drop ($-9.4$~pp on open-domain) and hurts every category; removing IDF \emph{improves} open-domain by $+3.1$~pp because square-root IDF zeroes hub nodes (speaker names) that open-domain preference questions actually benefit from.

\section{Conclusion}

We presented \method{}, a framework with no generative indexing calls that replaces generative indexing with a deterministic discriminative-parser pipeline over a mention incidence graph, addressing cost, non-determinism, and provenance loss in one move. On \locomo{}, it exceeds the strongest matched-protocol baselines at zero generative indexing calls (three locally matched runs); on LongMemEval-S, it remains competitive overall and leads or ties on single-session question types among local retrieval comparisons. The substrate composes with dense or LLM-generated complements without changing the indexing path.

\section{Limitations}
\label{sec:limitations}

\method{} prioritizes deterministic, source-traceable access over richer semantic, temporal, or relational structure. Evaluation is on English conversational QA with \texttt{en\_core\_web\_trf}; non-English and code-switched settings need dedicated evaluation. Fixed retrieval budgets can omit evidence for multi-session and temporal reasoning, and graph operations depend on history size. Transfer to other interaction streams is left for future work.

\begin{ack}
We sincerely thank the anonymous reviewers for their insightful comments and constructive suggestions. This research is supported by the National Natural Science Foundation of China Project (No.62606430, No. 623B2086), Sponsored by CCF-GHFund (No. OF 2026005), Sponsored by CIPS-SMP-Zhipu Large Model Fund, Ant Group and Shanghai AI Lab. We acknowledge funding support from the Guangzhou Municipal–University Joint Research Program (Grant No. SL2025A03J4490).

This work is partially supported by the Research Travel Grant of the Base of Red Bird MPhil (RBM) at The Hong Kong University of Science and Technology (Guangzhou). We would like to thank our RBM Project Supervisor Landu Jiang for the academic support.
\end{ack}

\begingroup
\small
\bibliographystyle{plainnat}
\bibliography{custom}
\endgroup


\appendix
\let\origsection\section
\renewcommand{\section}[1]{\clearpage\origsection{#1}}

\section{Reproducibility Details}
\label{appendix:reproducibility}

\paragraph{Mention identification cascade.}
The deterministic progressive cascade in Section~\ref{sec:index} fires in trust order, and each stage claims only the token spans the earlier stages did not cover. The five stages are: (i) named-entity spans whose label is on a closed proper-noun whitelist (\texttt{PERSON}, \texttt{GPE}, \texttt{LOC}, \texttt{ORG}, \texttt{WORK\_OF\_ART}, dates); (ii) noun chunks; (iii) verb-headed dependency subtrees containing a content verb; (iv) closed POS patterns over uncovered tokens; (v) leftover content singletons (proper nouns, content nouns, content verbs).
For example, ``Calvin found a new puppy at the adoption agency'' yields three nodes: \texttt{calvin} (Stage~i, named entity), \texttt{new puppy} (Stage~ii, noun chunk), \texttt{adoption agency} (Stage~ii, noun chunk); the verb \emph{found} is recovered as a content singleton in Stage~v.

\paragraph{Conservative dense tie-breaking.}
The numerical tolerance in Section~\ref{sec:rank} is $10^{-6}$ in absolute graph score. Within a tie, episodes are reranked by $\max_{u \in e}\cos(\phi_{\mathrm{txt}}(q),\phi_{\mathrm{txt}}(u))$ between the query and any utterance of $e$; outside ties dense similarity is ignored. This preserves the mention-graph signal in the typical case and uses dense similarity strictly to break draws.

\paragraph{Reader context.}
The reader receives the original utterances of the selected top-$5$ episodes with speaker labels, image captions, and session timestamps; no nodes, normalized keys, or co-occurrence summaries are passed in. Hybrid baselines fuse BM25 and dense rankings via reciprocal-rank fusion~\citep{cormack2009rrf} ($k{=}60$). The matched top-$10$/top-$5$ retrieval/reader budget follows the matched-protocol convention~\citep{shen2026anchormem,fang2025lightmem,xu2026structmem}.

\paragraph{Models.}
The primary reader is GPT-4o-mini (\texttt{gpt-4o-mini-2024-07-18}) and the judge is \texttt{Qwen2.5-32B-Instruct} in the primary evaluation; the separate graph-guidance ablation uses GPT-4o-mini as judge; the judge uses the LLM-as-a-judge rubric prompt in Appendix~\ref{appendix:judge}, held fixed across methods. A Qwen2.5-32B-Instruct reader-substitution check is reported in Table~\ref{tab:qwen-main} (Appendix~\ref{appendix:generalization}). The mention extractor is \texttt{spaCy}~\citep{honnibal2020spacy} \texttt{en\_core\_web\_trf} (zero-shot, content verbs included), reused on both dialogue and question sides; mention and episode embeddings use \texttt{all-MiniLM-L6-v2}~\citep{reimers2019sbert}.

\paragraph{Trial provenance.}
The implementation stores intermediate artifacts per trial (extracted mention nodes, episode-to-node incidence, bridge candidates, scored episodes, query plans, reader inputs, reader outputs, judge outputs).
Original retrieval/component diagnostics correspond to trial \texttt{2026-05-05-024-locomo-spacy-no-meta} (full pipeline, source-aligned \texttt{spaCy} parser on the question side, reader prompt without parser-side metadata, ten \locomo{} conversations, $1{,}540$ four-category questions). The earlier LLM-parser configuration (trial \texttt{2026-05-05-004-locomo-paper-graph}) is reported as the \emph{w/ LLM query parser} ablation row in Table~\ref{tab:ablation}; the reader-bottleneck oracle row in Table~\ref{tab:reader-bottleneck} is taken from trial \texttt{2026-05-05-009-locomo-paper-graph-oracle}, which reuses the same retrieval cache and swaps only the reader input.

\paragraph{Updated run provenance and compute.}
The primary QA result is the mean over seeds 42--44 ($75.89\pm0.10\%$); seed 42/43 results are 75.78\%/75.91\%. The original 74.85\% result and the separate GPT-4o-mini-judge graph ablation (75.65\%) are not interchangeable. Paired comparisons use outputs from the same run and protocol. Peak RSS and persisted index size are $3.332\pm0.186$~GiB and $138.885\pm0.057$~MiB for MGMem, $3.200\pm0.048$/$61.669\pm0.181$ for AnchorMem, $3.980\pm0.288$/$361.484\pm0.601$ for EMem, and $1.327\pm0.141$/$38.874\pm0.055$ for LightMem. Mean MGMem phase times (s) are 16.30 initialization, 37.67 segmentation, 85.66 extraction, 55.57 normalization/backfill/graph/embedding, and 2.40 serialization; independent total wall time includes orchestration.

\paragraph{Local retrieval configurations.}
BM25 uses \texttt{BM25Okapi} on raw sessions ($k_1=1.5$, $b=0.75$, $\epsilon=0.25$), with lowercased Unicode regex tokenization and no stemming or stop-word removal. Dense exhaustively ranks L2-normalized MiniLM raw-episode vectors by cosine; Hybrid fuses BM25 sessions and Dense episodes with RRF ($k=60$). NaiveRAG uses non-overlapping five-turn chunks, retains remainders, speakers, and captions, and scores episodes by maximum overlapping-chunk cosine with chronological tie-breaking. Top-five item budgets do not imply equal token budgets.

\section{Reader-Context Sensitivity}
\label{sec:reader-bottleneck}

In the original single-run diagnostic, conditional on a fixed reader/judge, how much accuracy is the memory access step itself leaving on the table?
Table~\ref{tab:reader-bottleneck} sweeps the reader's input from \method{}'s retrieved episodes to gold evidence (oracle), to the entire originating session, to the entire dialogue history.
Oracle evidence lifts judge accuracy from $0.7485$ to $0.895$ ($+14.6$~pp), the headroom available to a stronger memory selector.
Feeding the full originating session reaches $0.815$, while feeding the entire dialogue history ($30$--$50$k tokens) collapses the judge to $15.1\%$; $55\%$ of temporal errors under Full Context are confident refusals (``No mention of X'') even when the gold event is present.
This is a long-context grounding failure, not a context-size limit, and confirms that organizing the index---not enlarging the context---is the right axis. A qualitative breakdown of residual reader errors is in Appendix~\ref{appendix:reader-failures}.

\begin{table}[!htbp]
\centering
\small
\setlength{\tabcolsep}{6pt}
\renewcommand{\arraystretch}{1.15}
\begin{tabularx}{\linewidth}{Xrr}
\toprule
\textbf{Reader context} & \textbf{Judge Acc.} & \textbf{F1} \\
\midrule
Top-$k$ retrieved episodes (\method{}) & 0.7485 & 0.526 \\
Gold evidence episodes (oracle)        & 0.895 & 0.646 \\
Full gold sessions                     & 0.815 & 0.577 \\
Full dialogue history                  & 0.151 & 0.109 \\
\bottomrule
\end{tabularx}
\caption{Reader-context sweep with reader (GPT-4o-mini) and judge (\texttt{Qwen2.5-32B-Instruct}) held fixed ($1{,}540$ four-cat QA).}
\label{tab:reader-bottleneck}
\end{table}

\section{Residual Reader Errors}
\label{appendix:reader-failures}

Across conversations, hit@$5$ is only weakly correlated with judge accuracy while hit@$1$ correlates strongly with token-level F1: the reader frequently sees the correct evidence among the top five but still fails to produce the exact benchmark answer.
Manual inspection of hit@$5$-but-wrong single-hop questions surfaces two recurrent failures---the reader paraphrases too much or adds details not present in the gold, and it conflates multiple retrieved episodes when several discuss the same subject---both supporting the separation of memory access and answer generation as separable research problems.

\section{Index Construction Details}
\label{appendix:index-construction}

\paragraph{Episode segmentation.}
Each session $S_t$ is segmented at embedding change points.
For every gap $i$ between two consecutive utterances we compute a cohesion drop
\begin{equation}
\label{eq:gap}
c_i \;=\; 1 - \cos\!\bigl(\bar\phi(S_t[i{-}w:i]),\, \bar\phi(S_t[i:i{+}w])\bigr),
\end{equation}
with $\bar\phi(\cdot)$ the mean utterance embedding over a $w$-utterance window ($w=3$).
A boundary is inserted at every local peak satisfying $c_i > 1.10\,\bar c$, with a minimum segment length of two utterances; both parameters were chosen on \texttt{conv-26}, also included in the full benchmark and excluded in the nine-conversation analysis.
The resulting episodes are short enough for precise mention-based selection yet long enough to preserve the local context that anchors a recurring reference to its surrounding turns.

\paragraph{Closed-world incidence completion.}
The mention-extraction cascade can miss literal occurrences of a known mention in some episodes (e.g., \texttt{adoption agency} extracted in $e_a$ but not in $e_b$).
For every $(v,e)$ with $v \notin T_e$ we scan the raw text and captions of $e$ with a word-boundary regex against any known surface form of $v$ and add the incidence on a literal match; the step modifies $\mathcal{R}$ and $\{T_e\}$ only and never introduces new nodes.
After completion, $\mathcal{E}_G$, document frequencies $\mathrm{df}_v = |\{e : v \in T_e\}|$, and node embeddings $\phi(v)$ are recomputed from the final $\{T_e\}$.

\section{Algorithms}
\label{appendix:algorithms}

We summarize the offline index construction (Algorithm~\ref{alg:index}) and the online episode selection (Algorithm~\ref{alg:retrieve}) that, together, instantiate the framework described in Section~\ref{sec:method}.
Algorithm~\ref{alg:index} produces $G=(\mathcal{V},\mathcal{E}_G,\mathcal{R},\phi)$ from a raw dialogue with zero generative LLM calls during indexing; Algorithm~\ref{alg:retrieve} consumes $G$ together with a parsed query plan and emits a query-type-conditioned list of original episodes for the reader.

\begin{algorithm}[!htbp]
\small
\caption{Offline mention-incidence index construction.}
\label{alg:index}
\begin{algorithmic}[1]
\Require Dialogue $\mathcal{D}=\{S_1,\ldots,S_T\}$, normalizer $\nu(\cdot)$, mention encoder $\phi_{\mathrm{txt}}(\cdot)$, change-point parameters $(w,\kappa)$, similarity floor parameters used downstream
\Ensure Memory graph $G=(\mathcal{V},\mathcal{E}_G,\mathcal{R},\phi)$ with statistics $\{\mathrm{df}_v\}$
\State $\mathcal{E} \gets \varnothing,\;\; \mathcal{V}\gets\varnothing,\;\; \mathcal{R}\gets\varnothing,\;\; \mathit{surf}\gets\{\}$
\For{$t = 1,\ldots,T$}
  \State Compute gap scores $\{c_i\}$ for $S_t$ via Eq.~\eqref{eq:gap}; insert boundaries at local peaks with $c_i>\kappa\bar c$
  \State $\mathcal{E}\gets \mathcal{E}\cup\textsc{Episodes}(S_t)$
\EndFor
\For{$e \in \mathcal{E}$}
  \State $C_e \gets \textsc{ProgressiveCascade}(e) \cup \{\text{speaker}(u): u\in\text{utterances}(e)\}$
  \State $T_e \gets \{\nu(s) : s \in C_e,\;\nu(s)\neq\varepsilon\}$
  \For{$v \in T_e$}
    \State $\mathcal{V}\gets\mathcal{V}\cup\{v\}$;\;
           $\mathcal{R}\gets\mathcal{R}\cup\{(v,e)\}$;\;
           $\mathit{surf}[v]\gets\mathit{surf}[v]\cup\{s : \nu(s)=v\}$
  \EndFor
\EndFor
\For{$(v,e)$ with $v\in\mathcal{V},\,v\notin T_e$} \Comment{closed-world completion}
  \If{any pattern in $\mathit{surf}[v]\cup\{v\}$ matches the raw text or captions of $e$ under a word-boundary regex}
    \State $T_e\gets T_e\cup\{v\}$;\;\; $\mathcal{R}\gets\mathcal{R}\cup\{(v,e)\}$
  \EndIf
\EndFor
\State $\mathcal{E}_G \gets \{(\{u,v\}, w_{uv}) : w_{uv} = |\{e : u,v\in T_e\}| > 0\}$ \Comment{Eq.~\eqref{eq:cooccur}}
\State $\mathrm{df}_v \gets |\{e : v\in T_e\}|$ for every $v\in\mathcal{V}$
\State $\phi(v) \gets \phi_{\mathrm{txt}}(v)$ for every $v\in\mathcal{V}$ \Comment{single batch pass}
\State \Return $G=(\mathcal{V},\mathcal{E}_G,\mathcal{R},\phi)$, $\{\mathrm{df}_v\}$
\end{algorithmic}
\end{algorithm}

\begin{algorithm}[!htbp]
\small
\caption{Online subject-conditioned episode selection.}
\label{alg:retrieve}
\begin{algorithmic}[1]
\Require Query $q$, graph $G$, parser $\pi(\cdot)$, normalizer $\nu(\cdot)$, query encoder $\phi_{\mathrm{txt}}(\cdot)$, similarity floor $\theta$, ego budget $k_{\mathrm{ego}}$, total budget $k$, score-tie tolerance $\varepsilon$, reader budget $k_r$
\Ensure Ordered episode list $\mathcal{C}_q$
\State $(M_q, A_q) \gets \pi(q)$ \Comment{deterministic; same extractor as indexing}
\State $A_q^\star \gets \{\nu(a) : a \in A_q\} \cap \mathcal{V}$
\State $N \gets \bigl(\bigcup_{a\in A_q^\star}\{v : w_{av}>0\}\bigr)\setminus A_q^\star$ \Comment{Eq.~\eqref{eq:ego}}
\State $\mathcal{Z}_q \gets \varnothing$;\;\; $\textit{resolved}\gets\varnothing$
\For{$m \in M_q$ with $\nu(m)\in\mathcal{V}$} \Comment{Stage 1: exact}
  \State $\mathcal{Z}_q\gets \mathcal{Z}_q\cup\{(\nu(m),\,1.0,\,\textsc{exact})\}$;\;\; $\textit{resolved}\gets\textit{resolved}\cup\{m\}$
\EndFor
\For{$m \in M_q\setminus\textit{resolved}$}
  \State $S_m \gets \{(v,\cos(\phi_{\mathrm{txt}}(m),\phi(v))) : v\in\mathcal{V},\;\cos\geq\theta\}$
  \State Take top-$k_{\mathrm{ego}}$ by similarity from $S_m\cap (N\times\mathbb{R})$ as \textsc{subject\_ego\_bridge} \Comment{Stage 2}
  \State Fill remaining quota up to $k$ from $S_m\setminus(N\times\mathbb{R})$ as \textsc{semantic\_bridge} \Comment{Stage 3}
  \State Add accepted candidates with $\omega=\cos$ to $\mathcal{Z}_q$
\EndFor
\State $Z_q^\mathcal{V} \gets \{v : (v,\cdot,\cdot)\in\mathcal{Z}_q\}$
\For{$e \in \mathcal{E}$ with $H(e,q) = T_e\cap Z_q^\mathcal{V}\neq\varnothing$}
  \State Compute $S(e,q)$ from Eq.~\eqref{eq:score} using Eqs.~\eqref{eq:s-evi} and~\eqref{eq:s-cov-coh}
\EndFor
\State Sort episodes by $S(e,q)$ descending; for $e,e'$ with $|S(e,q)-S(e',q)|<\varepsilon$, resort by $\max_{u\in e}\cos(\phi_{\mathrm{txt}}(q),\phi_{\mathrm{txt}}(u))$
\State \Return the sorted list capped at the reader budget $k_r$
\end{algorithmic}
\end{algorithm}

\section{Additional Experimental Results}
\label{appendix:additional}

\begin{figure}[!htbp]
\centering
\includegraphics[width=0.95\linewidth]{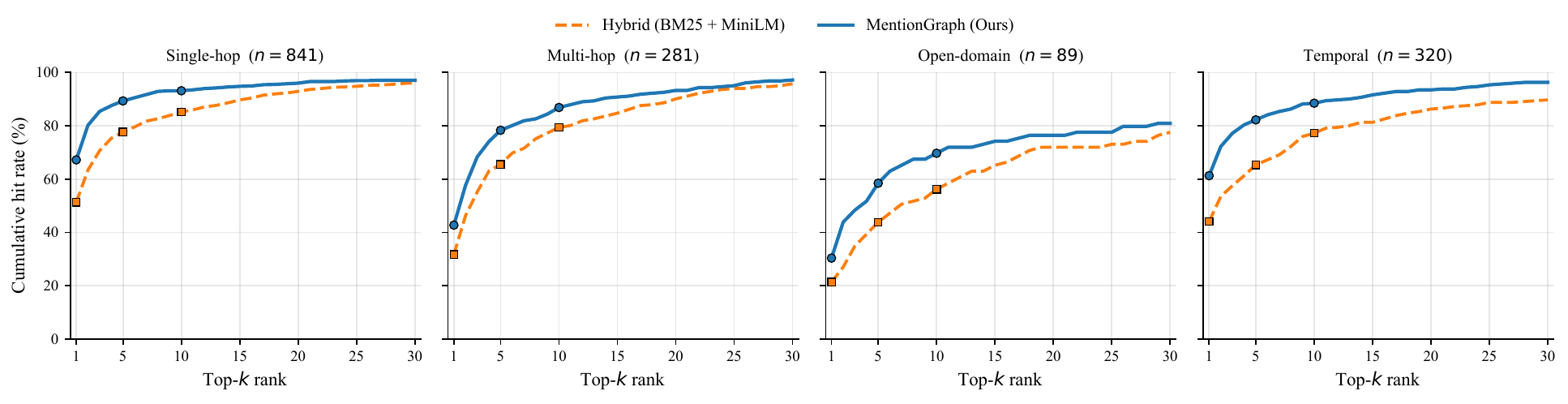}
\caption{Cumulative episode-level hit rate as a function of retrieval rank, broken down by question category. \method{} (blue, solid) and Hybrid (orange, dashed) under the same reader / judge configuration.}
\label{fig:cumulative-hit}
\end{figure}

Table~\ref{tab:retrieval-topk} reports per-category episode recovery hit@$k$ for $k\in\{1,3,5,10\}$. Figure~\ref{fig:cumulative-hit} plots the same retrieval breakdown against rank for \method{} and the Hybrid retrieval baseline; \method{} leads the Hybrid baseline across the reported ranks and categories.

\begin{table}[!htbp]
\centering
\small
\begin{tabular}{lrrrrr}
\toprule
Category & $n$ & hit@1 & hit@3 & hit@5 & hit@10 \\
\midrule
Multi-hop   & 281 & 0.427 & 0.683 & 0.781 & 0.866 \\
Temporal    & 320 & 0.613 & 0.770 & 0.820 & 0.883 \\
Open-domain &  89 & 0.292 & 0.449 & 0.539 & 0.640 \\
Single-hop  & 841 & 0.672 & 0.854 & 0.893 & 0.931 \\
\midrule
Overall     & 1{,}531 & 0.591 & 0.781 & 0.837 & 0.893 \\
\bottomrule
\end{tabular}
\caption{Per-category episode recovery hit@$k$ on \locomo{}.}
\label{tab:retrieval-topk}
\end{table}

\paragraph{Repeated runs and development exclusion.}
\begin{table}[!htbp]
\centering\small
\begin{tabular}{lccc}
\toprule
Method & All ten conversations & \texttt{conv-26} & Excluding \texttt{conv-26} \\
\midrule
MGMem & $75.89\pm0.10$ & $77.19\pm1.90$ & $75.74\pm0.25$ \\
EMem & $73.29\pm0.10$ & $74.78\pm1.52$ & $73.13\pm0.22$ \\
AnchorMem & $71.69\pm0.36$ & $76.97\pm1.14$ & $71.11\pm0.50$ \\
LightMem & $62.42\pm0.10$ & $64.47\pm1.74$ & $62.20\pm0.15$ \\
\bottomrule
\end{tabular}
\caption{Accuracy (\%), mean $\pm$ sample SD over seeds 42--44. Excluding the development conversation leaves 1,388 questions.}
\label{tab:dev-exclusion}
\end{table}
For the paired MGMem--EMem interval, 10,000 replicates resample ten conversation IDs with replacement, retain all question pairs within each conversation, and compute question-count-weighted differences. The same sampled conversation list is applied to all seeds before averaging; the percentile 95\% interval is $[+0.22,+5.15]$~pp. Only ten clusters are available. F1/BLEU-1 means (sample SD) are 0.3988 (0.0006)/0.3365 (0.0006) for MGMem, 0.3955 (0.0005)/0.3263 (0.0011) for EMem, 0.3588 (0.0026)/0.2920 (0.0030) for AnchorMem, and 0.3434 (0.0028)/0.2713 (0.0015) for LightMem.

\paragraph{Human audit and adversarial questions.}
Two annotators independently check 50 answers per answerable category and agree on final labels. Human--judge agreement is 46/50 multi-hop, 48/50 temporal, 46/50 open-domain, and 49/50 single-hop (189/200 overall; 94.50\%, Wilson 95\% CI [90.42\%, 96.90\%]). Evidence support counts are 46, 48, 45, and 49 (188/200; 94.00\%, CI [89.81\%, 96.53\%]). These are category-balanced sample rates. In matched Run 42 over 446 adversarial questions, correct refusals are 153 for MGMem (34.30\%, CI [30.05\%, 38.83\%]), 144 for Hybrid (32.29\%, [28.11\%, 36.76\%]), and 123 for AnchorMem (27.58\%, [23.64\%, 31.90\%]).

\section{Reader and Benchmark Generalization}
\label{appendix:generalization}

This appendix retains the original reader-substitution and LongMemEval-S evaluations. The Qwen reader also serves as judge, so this check is not judge-independent; it is separate from the revised three-run main comparison. LongMemEval-S generated-memory rows are quoted references rather than locally matched comparisons.

\begin{table*}[!htbp]
\centering
\small
\setlength{\tabcolsep}{8pt}
\renewcommand{\arraystretch}{1.15}
\begin{tabular}{@{}lccccc@{}}
\toprule
\textbf{Method} & \textbf{Single-Hop} & \textbf{Multi-Hop} & \textbf{Open-Domain} & \textbf{Temporal} & \textbf{Overall} \\
\midrule
\multicolumn{6}{l}{\textit{Retrieval baselines}} \\
\quad BM25                           & 71.62 & 37.50 & 31.91 & 55.17 & 59.43 \\
\quad Dense$^{\dagger}$              & 66.30 & 53.41 & \textbf{48.96} & 49.22 & 59.26 \\
\quad Hybrid$^{\ddagger}$            & 76.57 & 47.12 & 43.16 & 55.06 & 64.60 \\
\quad NaiveRAG                       & 62.32 & 42.29 & 40.00 & 46.20 & 53.89 \\
\quad HippoRAG~2                     & 62.66 & 56.38 & 42.71 & 56.07 & 58.90 \\
\quad Full Context                   & 12.35 & 15.64 & 21.05 & 9.52  & 12.90 \\
\midrule
\multicolumn{6}{l}{\textit{Generated-memory systems}} \\
\quad A-Mem                          & 47.56 & 43.62 & 28.12 & 34.58 & 42.92 \\
\quad Mem0                           & 36.75 & 29.43 & 31.25 & 31.15 & 33.88 \\
\quad LightMem                       & 64.33 & 50.71 & 40.63 & 61.99 & 59.87 \\
\quad MemoryOS                       & 55.65 & 56.74 & 42.71 & 23.36 & 48.31 \\
\quad AnchorMem                      & \underline{78.83} & \underline{58.51} & \underline{44.79} & \underline{62.31} & \underline{69.54} \\
\midrule
\rowcolor{black!4}
Ours                                 & \textbf{84.77} & \textbf{64.64} & 44.68 & \textbf{67.81} & \textbf{75.07} \\
\bottomrule
\end{tabular}
\caption{End-to-end QA judge accuracy on \locomo{} (\%) with \texttt{Qwen2.5-32B-Instruct} as the reader; the judge is also \texttt{Qwen2.5-32B-Instruct}, unchanged from Table~\ref{tab:main-qa}. This is the original single-run reader substitution, with the original segmentation and retrieval budgets; it is separate from the revised three-run main table. Generated-memory baselines and HippoRAG~2 numbers are quoted from~\citet{shen2026anchormem}, which evaluates each system under the same Qwen2.5-32B-Instruct reader+judge protocol. Best in \textbf{bold}, second-best \underline{underlined} among reported rows. $^{\dagger}$\,\texttt{all-MiniLM-L6-v2} encoder; $^{\ddagger}$\,BM25 fused with Dense$^{\dagger}$ via reciprocal-rank fusion ($k{=}60$).}
\label{tab:qwen-main}
\end{table*}

\begin{table*}[!htbp]
\centering
\footnotesize
\setlength{\tabcolsep}{3pt}
\renewcommand{\arraystretch}{1.10}
\begin{tabular}{@{}lcccccc|c@{}}
\toprule
\textbf{Method}
& \makecell{\textbf{Single-User}\\\scriptsize ($n{=}70$)}
& \makecell{\textbf{Single-Asst.}\\\scriptsize ($n{=}56$)}
& \makecell{\textbf{Single-Pref.}\\\scriptsize ($n{=}30$)}
& \makecell{\textbf{Multi-Session}\\\scriptsize ($n{=}133$)}
& \makecell{\textbf{Temporal}\\\scriptsize ($n{=}133$)}
& \makecell{\textbf{Knowl.-Upd.}\\\scriptsize ($n{=}78$)}
& \textbf{Overall} \\
\midrule
\multicolumn{8}{l}{\textit{Retrieval baselines}} \\
\quad BM25                  & \underline{95.71} & \underline{98.21} & \textbf{80.00} & \underline{59.40} & \textbf{67.67} & 74.36 & \textbf{74.60} \\
\quad Dense$^{\dagger}$     & 82.86 & \underline{98.21} & 63.33 & \underline{59.40} & 56.39 & 66.67 & 67.60 \\
\quad Hybrid$^{\ddagger}$   & \underline{95.71} & 96.43 & \textbf{80.00} & \textbf{65.41} & 60.90 & 71.79 & \underline{73.80} \\
\quad NaiveRAG              & 90.00 & 98.21 & 53.33 & 48.48 & 39.85 & 67.95 & 61.00 \\
\quad Full Context          & 87.14 & 89.29 & 36.67 & 45.45 & 31.58 & \textbf{76.92} & 56.80 \\
\midrule
\multicolumn{8}{l}{\textit{Generated-memory systems}} \\
\quad A-Mem                 & 92.86 & 96.43 & 46.67 & 48.87 & 47.36 & 64.11 & 62.60 \\
\quad Mem0                  & 81.43 & 41.07 & 60.00 & 46.21 & 40.15 & 70.12 & 53.61 \\
\quad LightMem              & 87.14 & 32.14 & 68.18 & 71.74 & 67.18 & 83.12 & 68.64 \\
\quad MemoryOS              & 80.00 & 64.29 & 30.00 & 31.06 & 32.33 & 48.72 & 44.80 \\
\midrule
\rowcolor{black!4}
Ours                        & \textbf{97.14} & \textbf{100.00} & \textbf{80.00} & 55.64 & \underline{61.65} & \underline{75.64} & 72.60 \\
\bottomrule
\end{tabular}
\caption{End-to-end QA judge accuracy on \textsc{LongMemEval-S}~\citep{wu2025longmemeval} (\%), broken down by the six question types. Generated-memory rows are quoted references; highlighting compares local retrieval rows only.}
\label{tab:longmemeval}
\end{table*}

\paragraph{Multi-session and temporal coverage.}
\begin{table}[!htbp]
\centering\small
\begin{tabular}{llccc}
\toprule
Type ($n=133$ each) & Method & Any gold session & All gold sessions & Conditional accuracy \\
\midrule
Multi-session & MGMem & 98.50\% & 65.41\% & 73.56\% \\
& BM25 & 96.99\% & 66.17\% & 82.95\% \\
& Hybrid & 98.50\% & 78.20\% & 79.81\% \\
Temporal & MGMem & 91.73\% & 68.42\% & 76.92\% \\
& BM25 & 94.74\% & 77.44\% & 76.70\% \\
& Hybrid & 96.24\% & 77.44\% & 68.93\% \\
\bottomrule
\end{tabular}
\caption{Top-five evidence coverage. Conditional accuracy is evaluated on each method's all-gold-session subset; subsets can differ.}
\label{tab:lme-coverage}
\end{table}
Incomplete all-session coverage is one measured factor, especially for temporal questions, and does not explain the full multi-session gap. Conditional accuracies are not a controlled cross-method comparison. The paired question-bootstrap interval for MGMem minus BM25 overall ($-2.00$~pp) is $[-5.40,+1.40]$~pp, which is not evidence of equivalence.

\section{Additional Ablations}
\label{appendix:additional-ablations}

\paragraph{Component ablation: full retrieval metrics.}
Table~\ref{tab:ablation} reports the per-ablation hit@1, hit@5, MRR, and absolute / relative drops in judge accuracy that underlie the heatmap in Figure~\ref{fig:strat-ablation}\,(b).

\begin{table*}[!htbp]
\centering
\scriptsize
\setlength{\tabcolsep}{3pt}
\renewcommand{\arraystretch}{1.15}
\begin{tabularx}{\textwidth}{@{}p{0.20\textwidth}Xcccccc@{}}
\toprule
\textbf{Variant} & \textbf{Hypothesis tested} & hit@1 & hit@5 & MRR & Judge Acc. & $\Delta$\,pp & $\Delta$\,\% \\
\midrule
\rowcolor{black!4}
\textbf{Full \method{}} & all mechanisms & \textbf{0.591} & \textbf{0.837} & \textbf{0.701} & \textbf{0.7485} & \textbf{0.00} & --- \\
\midrule
w/ LLM query parser      & query mentions from separate LLM parser & 0.574 & 0.829 & 0.689 & 0.723 & $-2.6$ & $-3.4$ \\
Exact match only         & flat literal mention-IDF access & 0.533 & 0.790 & 0.651 & 0.689 & $-6.0$ & $-7.9$ \\
w/o IDF                  & remove rare-mention weighting & 0.559 & 0.802 & 0.669 & 0.708 & $-4.1$ & $-5.4$ \\
w/o subject-local bridge & remove subject ego expansion & 0.534 & 0.791 & 0.652 & 0.688 & $-6.1$ & $-8.1$ \\
w/o graph coherence      & remove co-occurrence tie-shaper & 0.574 & 0.830 & 0.689 & 0.727 & $-2.2$ & $-2.9$ \\
w/o conservative dense   & remove dense tie-break & 0.548 & 0.811 & 0.664 & 0.706 & $-4.3$ & $-5.7$ \\
\midrule
\multicolumn{8}{l}{\textit{Stage-isolation ablation}} \\
\quad Stage 1+2, no Stage 3 & disable full-graph bridge stage only & 0.589 & 0.833 & 0.698 & --- & --- & --- \\
\bottomrule
\end{tabularx}
\caption{Original single-run component ablation on \locomo{} ($1{,}540$ four-cat QA), separate from the revised main table. Each row removes or replaces one mechanism from the full method. $\Delta$\,pp is the absolute drop in judge accuracy in percentage points; $\Delta$\,\% is the same drop as a fraction of Full \method{}'s judge accuracy. The \emph{Stage~1+2} row is retrieval-only (judge skipped).}
\label{tab:ablation}
\end{table*}

\paragraph{Graph guidance, Stage 3, and segmentation.}
A separate matched GPT-4o-mini-reader/judge run yields 75.65\% for Full and 73.31\% for Inverted-Index-Only ($+2.34$~pp); hit@1/hit@5/MRR are 0.5911/0.8372/0.7010 versus 0.5651/0.8008/0.6723. The flat variant retains mentions, embeddings, incidence, global bridging, IDF, and coverage, removing subject-local guidance and coherence. This measures graph-guided retrieval within the architecture; no paired CI is available, so the result is descriptive.
A separate 1,536-question retrieval run finds Stage~3 adds 0.26/0.39/0.31~pp in hit@1/hit@5/MRR overall, 1.33/2.00/1.58~pp on 300 queries where it produces candidates, and 2.39/3.35/2.69~pp on 209 queries without a subject anchor.
\begin{table}[!htbp]
\centering\small
\begin{tabular}{lrrrr}
\toprule
Segmentation & hit@1 & hit@5 & MRR & Words@5 \\
\midrule
Session & 0.6335 & 0.8958 & 0.7473 & 3200.3 \\
Topic-fine ($\kappa=0.8$) & 0.5534 & 0.8092 & 0.6680 & 735.9 \\
Topic-default ($\kappa=1.1$) & 0.5924 & 0.8379 & 0.7018 & 993.0 \\
Topic-coarse ($\kappa=1.4$) & 0.6250 & 0.8763 & 0.7340 & 2108.1 \\
\bottomrule
\end{tabular}
\caption{Separate 1,536-question segmentation analysis: coarser units improve recall while returning more text. Absolute metrics are not pooled with the original 1,531-question retrieval analysis.}
\label{tab:segmentation}
\end{table}

\paragraph{IDF and combiner.}
Tables~\ref{tab:idf} and~\ref{tab:combiner} report two ablations on development conversation \texttt{conv-26} (199 questions) along axes complementary to those in Section~\ref{sec:analysis}: the IDF variant used in $S_{\mathrm{evi}}$ (Section~\ref{sec:rank}), and the dense-fusion strategy applied on top of the mention-graph ranking.

\begin{table}[!htbp]
\centering
\small
\begin{tabular}{lrrrr}
\toprule
IDF mode & hit@1 & hit@5 & Cat3 h@1 & Cat4 h@1 \\
\midrule
Standard & 0.638 & 0.878 & 0.455 & 0.729 \\
Smoothed & 0.633 & 0.883 & 0.364 & 0.743 \\
Sqrt & 0.628 & 0.888 & 0.455 & 0.729 \\
No IDF & 0.587 & 0.837 & 0.273 & 0.714 \\
\bottomrule
\end{tabular}
\caption{IDF ablation on a 199-question development conversation. These runs predate the spaCy-parser switch (LLM parser used to keep cross-mode comparison clean). Production uses square-root IDF.}
\label{tab:idf}
\end{table}

\begin{table}[!htbp]
\centering
\small
\begin{tabular}{lrr}
\toprule
Combiner & hit@1 & Cat4 hit@1 \\
\midrule
Tie-break only & 0.638 & 0.729 \\
RRF + dense & 0.480 & 0.586 \\
Dense access top-10 & 0.398 & 0.514 \\
\bottomrule
\end{tabular}
\caption{Dense-fusion ablation on the same dev conversation. Production uses tie-break only.}
\label{tab:combiner}
\end{table}

\section{Hyperparameter Sensitivity}
\label{appendix:hyperparameters}

Production values were fixed on \texttt{conv-26} beforehand (Section~\ref{sec:settings}); this sweep is a post-hoc sensitivity analysis and was not used for selection.

\begin{figure}[!htbp]
\centering
\includegraphics[width=\linewidth]{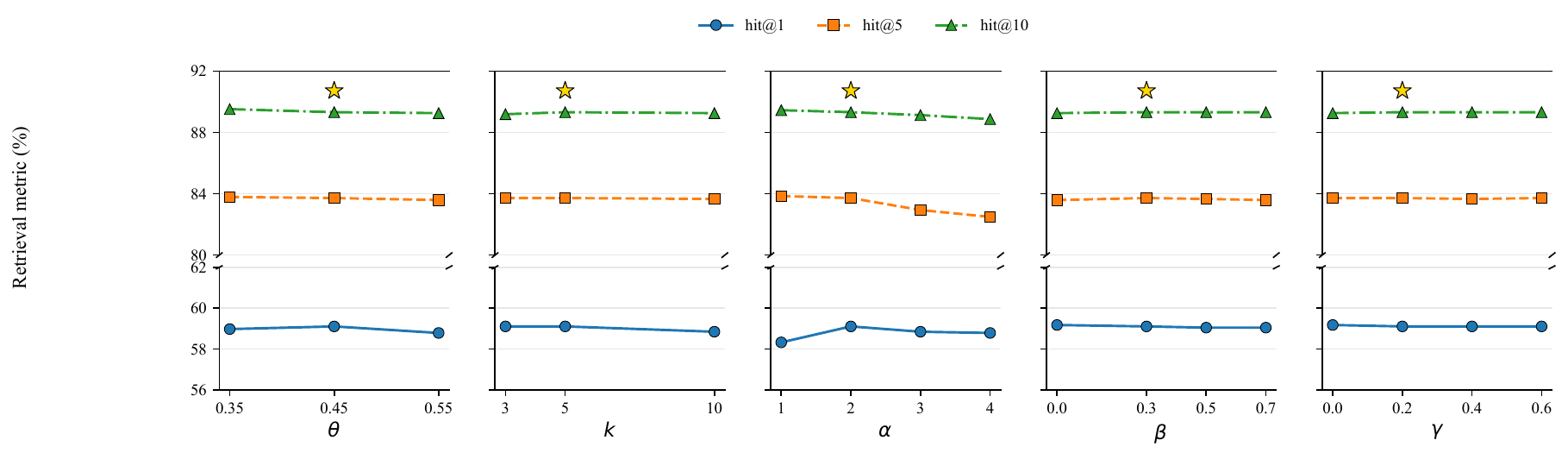}
\caption{Retrieval-metric sensitivity to all five \method{} hyperparameters: bridge similarity floor $\theta$, evidence-node budget $k$, evidence-aggregation exponent $\alpha$, cue-coverage weight $\beta$, pair-coherence weight $\gamma$. Production setting marked with $\star$.}
\label{fig:sensitivity}
\end{figure}

Table~\ref{tab:hyperparameters} and Figure~\ref{fig:sensitivity} report sensitivity sweeps over the five \method{} hyperparameters: the bridge similarity floor $\theta$ (production: $0.45$), the per-mention evidence-node budget $k$ (production: $5$), the evidence-aggregation exponent $\alpha$ (production: $2$), the cue-coverage weight $\beta$ (production: $0.3$), and the pair-coherence weight $\gamma$ (production: $0.2$).
Each sweep varies one parameter and holds the others at their production values; all four retrieval metrics are reported.
Aggregating across all $18$ tested settings, hit@1 spans $\leq 0.85$~pp and hit@5 spans $\leq 1.36$~pp of the production setting, with the $\alpha$ sweep contributing the largest single-axis variation and $\beta,\gamma$ being effectively flat.
This supports the deterministic, hyperparameter-light design: the production setting is robust within the practical neighborhood, and no axis is fragile.

\begin{table}[!htbp]
\centering
\small
\setlength{\tabcolsep}{5pt}
\begin{tabular}{llcccc}
\toprule
Sweep axis & Setting & hit@1 & hit@5 & hit@10 & MRR \\
\midrule
\multirow{3}{*}{$\theta$ (\textit{prod 0.45})}
              & $\theta=0.35$        & 0.5898 & 0.8379 & 0.8952 & 0.7011 \\
              & $\theta=0.45$ (prod) & 0.5911 & 0.8372 & 0.8932 & 0.7010 \\
              & $\theta=0.55$        & 0.5879 & 0.8359 & 0.8926 & 0.6998 \\
\midrule
\multirow{3}{*}{$k$ (\textit{prod 5})}
              & $k=3$         & 0.5911 & 0.8372 & 0.8919 & 0.7014 \\
              & $k=5$ (prod)  & 0.5911 & 0.8372 & 0.8932 & 0.7010 \\
              & $k=10$        & 0.5885 & 0.8366 & 0.8926 & 0.6995 \\
\midrule
\multirow{4}{*}{$\alpha$ (\textit{prod 2})}
              & $\alpha=1$        & 0.5833 & 0.8385 & 0.8945 & 0.6965 \\
              & $\alpha=2$ (prod) & 0.5911 & 0.8372 & 0.8932 & 0.7010 \\
              & $\alpha=3$        & 0.5885 & 0.8294 & 0.8913 & 0.6980 \\
              & $\alpha=4$        & 0.5879 & 0.8249 & 0.8887 & 0.6957 \\
\midrule
\multirow{4}{*}{$\beta$ (\textit{prod 0.3})}
              & $\beta=0.0$         & 0.5918 & 0.8359 & 0.8926 & 0.7007 \\
              & $\beta=0.3$ (prod)  & 0.5911 & 0.8372 & 0.8932 & 0.7010 \\
              & $\beta=0.5$         & 0.5905 & 0.8366 & 0.8932 & 0.7007 \\
              & $\beta=0.7$         & 0.5905 & 0.8359 & 0.8932 & 0.7007 \\
\midrule
\multirow{4}{*}{$\gamma$ (\textit{prod 0.2})}
              & $\gamma=0.0$        & 0.5918 & 0.8372 & 0.8926 & 0.7014 \\
              & $\gamma=0.2$ (prod) & 0.5911 & 0.8372 & 0.8932 & 0.7010 \\
              & $\gamma=0.4$        & 0.5911 & 0.8366 & 0.8932 & 0.7007 \\
              & $\gamma=0.6$        & 0.5911 & 0.8372 & 0.8932 & 0.7007 \\
\bottomrule
\end{tabular}
\caption{Five-axis sensitivity sweep on \locomo{} ($1{,}540$ four-cat QA, retrieval-only). Each block varies one hyperparameter while holding the others at production (marked \emph{prod}).}
\label{tab:hyperparameters}
\end{table}

\section{Graph Statistics}
\label{appendix:graph-statistics}

Table~\ref{tab:graph-statistics} reports structural statistics of the mention incidence graph across the ten \locomo{} conversations.
Table~\ref{tab:resolution-paths} reports the share of queries whose evidence is supplied by each of the three resolver stages of Section~\ref{sec:resolve}.

\begin{table}[!htbp]
\centering
\small
\begin{tabular}{lrrrr}
\toprule
Statistic & Mean & Std. & Min & Max \\
\midrule
episodes per dialogue   & 120.1   & 23.1   & 78    & 138    \\
mention nodes           & 1{,}608.9 & 260.9  & 1{,}039 & 1{,}927 \\
incidence links         & 6{,}458.4 & 1{,}231.9 & 3{,}888 & 7{,}828 \\
co-occurrence edges     & 111{,}762 & 22{,}078 & 65{,}164 & 146{,}295 \\
avg. node degree        & 138.32  & 9.78   & 125.44 & 155.80 \\
\bottomrule
\end{tabular}
\caption{Structural statistics of the mention incidence graph across the ten \locomo{} conversations.}
\label{tab:graph-statistics}
\end{table}

\begin{table}[!htbp]
\centering
\small
\begin{tabular}{lrr}
\toprule
Resolution path & Count & Share \\
\midrule
Exact match only                       & 663 & 43.1\% \\
Exact + subject-local bridge           & 634 & 41.2\% \\
Full-graph bridge (semantic + whole-q) & 221 & 14.4\% \\
Subject-local bridge only (no exact)   &  22 &  1.4\% \\
\bottomrule
\end{tabular}
\caption{Resolver-stage usage across the 1{,}540 \locomo{} four-cat queries.}
\label{tab:resolution-paths}
\end{table}




\paragraph{Complexity.}
With $N$ episodes, $M$ mention nodes, $r_e$ mentions in episode $e$, $I=\sum_e r_e$, and embedding dimension $d$, co-occurrence construction costs $O(\sum_e r_e^2)$, storage $O(I+|\mathcal{E}_G|+Md)$, full semantic scanning $O(|M_q|Md)$ per query, and subject-neighbor inspection $O(\sum_a\deg(a))$. Literal completion uses up to $O(MN)$ match attempts plus text scanning. Stage~2 has no recursive expansion; five accepted nodes per mention and five reader episodes bound outputs, not graph-size-dependent computation.

\section{Paradigm Comparison}

Table~\ref{tab:paradigm-comparison} contrasts \method{} with the dominant long-term memory paradigms along five dimensions: the unit stored in memory, the structure imposed on those units, whether dialogue history is rewritten during construction, the retrieval signal used at query time, and the provenance of the content fed to the reader.

\begin{table*}[!htbp]
\centering
\small
\begin{tabularx}{\textwidth}{lXXXXX}
\toprule
Paradigm & Memory unit & Structure & Rewrites history? & Retrieval signal & Provenance \\
\midrule
Flat RAG & chunks & none / vector index & no & dense or sparse similarity & raw chunks \\
Rewriting memory & summaries / notes & evolving memory store & yes & generated memory text & often indirect \\
Fact or semantic graph memory & facts / entities & generated relations & partly & graph traversal and similarity & fact-to-source links vary \\
Hierarchical memory & topics / episodes / facts & hierarchy or hypergraph & partly & coarse-to-fine retrieval & source context varies \\
\method{} & raw episodes indexed by mentions & mention incidence graph & no & mention coverage and graph coherence & raw episodes \\
\bottomrule
\end{tabularx}
\caption{Memory paradigms compared along five dimensions: memory unit, structure, whether history is rewritten, retrieval signal, and reader provenance.}
\label{tab:paradigm-comparison}
\end{table*}

\section{Prompts}
\label{appendix:judge}

This appendix lists the two LLM prompts used at evaluation time.
There is no LLM-side query-parsing prompt in the production configuration: the question is parsed by running the same dialogue spaCy mention extractor on the question text, so $(M_q, A_q)$ are produced deterministically without any generative LLM call.

\subsection{Reader Prompt}

The reader is called with a system message containing general instructions for grounded multi-session QA, followed by a per-question user message that pastes the question, the parsed query type and subject set, and the formatted episode evidence.
The system message is held fixed across all matched-protocol runs.

\begin{promptbox}{Reader System Prompt}
\small
You are answering questions about a long-running multi-session dialogue.
You have access to a set of memories --- each memory is a snippet of dialogue from one session, prefixed with a session\_time timestamp.
Your answer must be grounded in these memories; you may use general world knowledge only to interpret or complete information that the memories themselves introduce (e.g., resolving a landmark named by description).

\smallskip
\textbf{Instructions.}

\begin{enumerate}[leftmargin=*,itemsep=2pt,topsep=2pt]
\item Carefully analyze every memory. Synthesize information across multiple memories when a single one is insufficient.
\item Pay close attention to session\_time. If memories contradict, the most recent memory is the source of truth.
\item For factual questions, look for direct evidence in the memories.
\item For time questions, you MUST resolve relative phrases against the memory's session\_time:
  \begin{itemize}[leftmargin=*,itemsep=1pt,topsep=1pt]
  \item Precise (``yesterday'', ``2 days ago''): compute the exact date (e.g., ``yesterday'' + session\_time 5 July 2023 $\rightarrow$ ``4 July 2023'').
  \item Vague (``last week'', ``last summer''): anchor to the reference date (e.g., ``last week'' + session\_time 10 October 2023 $\rightarrow$ ``the week before 10 October 2023'').
  \end{itemize}
  Always convert relative references to absolute dates, months, or years in the final answer.
\item Match the LoCoMo gold-answer style for date precision: Year only ``2022''; Month + year ``July 2023''; Full date ``2 July 2023''; Anchored relative ``the Friday before 15 July 2023'' / ``the week of 23 August 2023''; Range start ``Since 2016''.
\item Do NOT confuse character names mentioned inside memory text with the actual speakers who own the dialogue.
\item Output ONLY the final answer:
  \begin{itemize}[leftmargin=*,itemsep=1pt,topsep=1pt]
  \item factoid / single-fact: under 5--6 words, no preamble.
  \item list / aggregation: items separated by commas, each item short.
  \item comparative: per-subject facts, separated by ``;''.
  \end{itemize}
  No reasoning, no quoting of memories, no ``Based on the memories\ldots''.
\end{enumerate}

\smallskip
\textbf{Approach} (think step by step internally; do NOT print the steps).

\begin{enumerate}[leftmargin=*,itemsep=2pt,topsep=2pt]
\item Identify the memories relevant to the question.
\item Synthesize across memories if needed; resolve conflicts by session\_time recency.
\item Extract explicit dates, names, events.
\item If a calculation is needed (relative $\rightarrow$ absolute date), perform it.
\item Formulate a precise, concise answer.
\item Verify the answer addresses the question and obeys the format rules above.
\end{enumerate}
\end{promptbox}

\smallskip

\begin{promptbox}{Reader User Message}
\small
Question: \{question\}

\smallskip
Memories:

\{formatted episodes: each prefixed by ``\#\#\# Episode \{eid\}'' and ``session\_time: \{date\}'', followed by speaker-tagged turns ``[\{dia\_id\}] \{speaker\}: \{text\}''\}

\smallskip
Final answer:
\end{promptbox}

\subsection{Judge Prompt}

The judge is invoked once per question with the entire rubric, the gold answer, and the model's predicted answer in a single user message. The model is asked to emit a one-sentence rationale followed by a JSON-wrapped CORRECT/WRONG label, which our parser strips back down to the binary outcome reported in the tables.
The same prompt is used for every method; only \texttt{\{question\}}, \texttt{\{gold\}}, and \texttt{\{pred\}} are filled in per example.

\begin{promptbox}{Prompt for LLM Judge}
\small
Your task is to label an answer to a question as `CORRECT' or `WRONG'. You will be given the following data:\\
(1) a question (posed by one user to another user),\\
(2) a `gold' (ground truth) answer,\\
(3) a generated answer\\
which you will score as CORRECT/WRONG.

\smallskip
The point of the question is to ask about something one user should know about the other user based on their prior conversations.

\smallskip
The gold answer will usually be a concise and short answer that includes the referenced topic, for example:\\
Question: Do you remember what I got the last time I went to Hawaii?\\
Gold answer: A shell necklace\\
The generated answer might be much longer, but you should be generous with your grading --- as long as it touches on the same topic as the gold answer, it should be counted as CORRECT.

\smallskip
For time related questions, the gold answer will be a specific date, month, year, etc.\ The generated answer might be much longer or use relative time references (like ``last Tuesday'' or ``next month''), but you should be generous with your grading --- as long as it refers to the same date or time period as the gold answer, it should be counted as CORRECT. Even if the format differs (e.g., ``May 7th'' vs ``7 May''), consider it CORRECT if it's the same date.

\smallskip
Now it's time for the real question:\\
Question: \{question\}\\
Gold answer: \{gold\_answer\}\\
Generated answer: \{generated\_answer\}

\smallskip
First, provide a short (one sentence) explanation of your reasoning, then finish with CORRECT or WRONG. Do NOT include both CORRECT and WRONG in your response, or it will break the evaluation script. Just return the label CORRECT or WRONG in a json format with the key as ``label''.
\end{promptbox}


\let\section\origsection
\newpage
\input{checklist_camera_ready.tex}

\end{document}

%% file: checklist_camera_ready.tex
\section*{NeurIPS Paper Checklist}

\begin{enumerate}

\item {\bf Claims}
    \item[] Question: Do the main claims made in the abstract and introduction accurately reflect the paper's contributions and scope?
    \item[] Answer: \answerYes{}
    \item[] Justification: The abstract and introduction scope the claims to the evaluated memory architecture; Table~\ref{tab:main-qa} reports matched reruns and Table~\ref{tab:efficiency} complete indexing costs. Benchmark and reader-substitution scope are specified in Appendix~\ref{appendix:generalization}.

\item {\bf Limitations}
    \item[] Question: Does the paper discuss the limitations of the work performed by the authors?
    \item[] Answer: \answerYes{}
    \item[] Justification: The Limitations section discusses English parsing, evidence coverage for multi-session and temporal reasoning, graph-size-dependent computation, and transfer to other interaction streams.

\item {\bf Theory assumptions and proofs}
    \item[] Question: For each theoretical result, does the paper provide the full set of assumptions and a complete (and correct) proof?
    \item[] Answer: \answerNA{}
    \item[] Justification: The paper is an empirical systems contribution; it does not state formal theorems and therefore has no proofs. All algorithmic equations (mention extraction, IDF weighting, scoring) are constructive and stated alongside the components they govern in Section~\ref{sec:method}.

    \item {\bf Experimental result reproducibility}
    \item[] Question: Does the paper fully disclose all the information needed to reproduce the main experimental results of the paper to the extent that it affects the main claims and/or conclusions of the paper (regardless of whether the code and data are provided or not)?
    \item[] Answer: \answerYes{}
    \item[] Justification: Section~\ref{sec:method} fully specifies the indexing pipeline; Section~\ref{sec:experiments} specifies datasets (\locomo{} four-category $1{,}540$-question subset; full LongMemEval-S $500$-question test set), the reader (GPT-4o-mini) and judge (\texttt{Qwen2.5-32B-Instruct}), and the matched-protocol convention; Appendix~\ref{appendix:reproducibility} lists trial identifiers; Appendix~\ref{appendix:index-construction} gives change-point and incidence-completion parameters; Appendix~\ref{appendix:algorithms} provides pseudocode for offline indexing and online retrieval; Appendix~\ref{appendix:judge} gives the verbatim reader and judge prompts.

\item {\bf Open access to data and code}
    \item[] Question: Does the paper provide open access to the data and code, with sufficient instructions to faithfully reproduce the main experimental results, as described in supplemental material?
    \item[] Answer: \answerYes{}
    \item[] Justification: An anonymous code repository is linked in the title footnote on page~1. The two evaluation datasets (\locomo{}~\citep{maharana2024locomo}, LongMemEval~\citep{wu2025longmemeval}) are public; we include preprocessing scripts, the deterministic indexing code, the matched-protocol baselines used here, and the reader/judge prompts in the released repository.

\item {\bf Experimental setting/details}
    \item[] Question: Does the paper specify all the training and test details (e.g., data splits, hyperparameters, how they were chosen, type of optimizer) necessary to understand the results?
    \item[] Answer: \answerYes{}
    \item[] Justification: Section~\ref{sec:experiments} specifies the data split (\locomo{} four-category $1{,}540$-question subset; full LongMemEval-S $500$-question test set), the reader (\texttt{gpt-4o-mini-2024-07-18}), the judge (\texttt{Qwen2.5-32B-Instruct}), the mention extractor (\texttt{spaCy} \texttt{en\_core\_web\_trf}), the embedding model (\texttt{all-MiniLM-L6-v2}), and the production hyperparameters ($\sqrt{}$-IDF, $\alpha{=}2$, $\beta{=}0.3$, $\gamma{=}0.2$, $\theta{=}0.45$); the framework has no trainable parameters. Hyperparameter sensitivity is reported in Appendix~\ref{appendix:hyperparameters}.

\item {\bf Experiment statistical significance}
    \item[] Question: Does the paper report error bars suitably and correctly defined or other appropriate information about the statistical significance of the experiments?
    \item[] Answer: \answerYes{}
    \item[] Justification: The local primary comparisons report mean and sample SD across seeds 42--44; Appendix~\ref{appendix:additional} specifies the paired conversation bootstrap and its ten-cluster limitation. Original single-run diagnostics are identified separately.

\item {\bf Experiments compute resources}
    \item[] Question: For each experiment, does the paper provide sufficient information on the computer resources (type of compute workers, memory, time of execution) needed to reproduce the experiments?
    \item[] Answer: \answerYes{}
    \item[] Justification: Table~\ref{tab:efficiency} reports complete indexing wall time and generative calls/tokens over three fresh-index runs on an Apple M5 Pro. Section~\ref{sec:efficiency} reports collection-level retrieval/reader wall totals; Appendix~\ref{appendix:reproducibility} reports footprint and phase timing.

\item {\bf Code of ethics}
    \item[] Question: Does the research conducted in the paper conform, in every respect, with the NeurIPS Code of Ethics \url{https://neurips.cc/public/EthicsGuidelines}?
    \item[] Answer: \answerYes{}
    \item[] Justification: We use only public benchmarks (\locomo{}, LongMemEval) under their stated terms; no human-subjects data was collected, no scraping was performed, and the released code does not depend on private or restricted-access assets. Anonymity is preserved throughout.

\item {\bf Broader impacts}
    \item[] Question: Does the paper discuss both potential positive societal impacts and negative societal impacts of the work performed?
    \item[] Answer: \answerYes{}
    \item[] Justification: Positive impacts: deterministic, provenance-preserving memory makes long-horizon agent behavior more auditable, reduces the energy cost of repeated index re-construction (zero generative LLM calls), and avoids storing or surfacing LLM-synthesized content that may misrepresent the original interaction. Negative impacts: the framework indexes user-specific recurring references, which are personally identifying by design; deployment should follow the user-data handling and consent norms of the host application. These trade-offs are discussed in the Limitations section.

\item {\bf Safeguards}
    \item[] Question: Does the paper describe safeguards that have been put in place for responsible release of data or models that have a high risk for misuse (e.g., pre-trained language models, image generators, or scraped datasets)?
    \item[] Answer: \answerNA{}
    \item[] Justification: The paper does not release any pre-trained language model, image generator, or scraped dataset. \method{} is an indexing/retrieval framework whose only released artifact is source code; the language and embedding models it depends on are unmodified third-party models.

\item {\bf Licenses for existing assets}
    \item[] Question: Are the creators or original owners of assets (e.g., code, data, models), used in the paper, properly credited and are the license and terms of use explicitly mentioned and properly respected?
    \item[] Answer: \answerYes{}
    \item[] Justification: We cite the original source of every asset used: \locomo{}~\citep{maharana2024locomo} and LongMemEval~\citep{wu2025longmemeval} for benchmarks; \texttt{spaCy}~\citep{honnibal2020spacy} (MIT) and the \texttt{en\_core\_web\_trf} pipeline for parsing; Sentence-BERT MiniLM~\citep{reimers2019sbert} (Apache 2.0) for embeddings; GPT-4o-mini for the reader and \texttt{Qwen2.5-32B-Instruct} for the primary judge; matched-protocol baseline numbers are attributed to~\citet{shen2026anchormem,fang2025lightmem,xu2026structmem}.

\item {\bf New assets}
    \item[] Question: Are new assets introduced in the paper well documented and is the documentation provided alongside the assets?
    \item[] Answer: \answerYes{}
    \item[] Justification: The new asset introduced is the \method{} reference implementation, released through the anonymous repository linked on page~1. The repository ships with a README describing the indexing pipeline, the configuration files for the production hyperparameters, the reader/judge prompts (also reproduced in Appendix~\ref{appendix:judge}), and scripts that reproduce the \locomo{} and LongMemEval results reported in the paper.

\item {\bf Crowdsourcing and research with human subjects}
    \item[] Question: For crowdsourcing experiments and research with human subjects, does the paper include the full text of instructions given to participants and screenshots, if applicable, as well as details about compensation (if any)?
    \item[] Answer: \answerNA{}
    \item[] Justification: An additional two-annotator audit of 200 benchmark answers checks correctness and evidence support (Appendix~\ref{appendix:additional}); no crowdsourcing or participant interaction study is conducted.

\item {\bf Institutional review board (IRB) approvals or equivalent for research with human subjects}
    \item[] Question: Does the paper describe potential risks incurred by study participants, whether such risks were disclosed to the subjects, and whether Institutional Review Board (IRB) approvals (or an equivalent approval/review based on the requirements of your country or institution) were obtained?
    \item[] Answer: \answerNA{}
    \item[] Justification: The paper does not involve human subjects, so no IRB approval was required.

\item {\bf Declaration of LLM usage}
    \item[] Question: Does the paper describe the usage of LLMs if it is an important, original, or non-standard component of the core methods in this research? Note that if the LLM is used only for writing, editing, or formatting purposes and does \emph{not} impact the core methodology, scientific rigor, or originality of the research, declaration is not required.
    \item[] Answer: \answerYes{}
    \item[] Justification: The core indexing method of \method{} contains no generative LLM call by design (this is one of the paper's central claims; see Section~\ref{sec:method}). Generative LLMs enter the evaluation pipeline only as the downstream reader (\texttt{gpt-4o-mini-2024-07-18}) and the LLM-as-a-judge (\texttt{Qwen2.5-32B-Instruct}); both roles are declared in Section~\ref{sec:experiments} and their full prompts are reproduced verbatim in Appendix~\ref{appendix:judge}, matching prior matched-protocol work.

\end{enumerate}